\documentclass[]{fairmeta}

\usepackage{graphicx}
\usepackage{multicol}
\usepackage{multirow}
\usepackage{booktabs}
\usepackage{times}
\usepackage{subcaption}
\usepackage{wrapfig}
\usepackage{arydshln}
\usepackage{xcolor}
\usepackage{amsmath}
\usepackage{amsfonts}
\definecolor{myred}{rgb}{1, 0.85, 0.85}
\definecolor{myblue}{rgb}{0.85, 0.9, 1}
\usepackage{soul}
\usepackage{makecell}

\title{Verifying Chain-of-Thought Reasoning via Its Computational Graph}

\author[1,3,*]{Zheng Zhao}
\author[1]{Yeskendir Koishekenov}
\author[2]{Xianjun Yang}
\author[1]{Naila Murray}
\author[1]{Nicola Cancedda}

\affiliation[1]{FAIR at Meta}
\affiliation[2]{Meta Superintelligence Labs}
\affiliation[3]{University of Edinburgh}

\contribution[*]{Work done during an internship at FAIR}

\abstract{
Current Chain-of-Thought (CoT) verification methods predict reasoning correctness based on outputs (black-box) or activations (gray-box), but offer limited insight into \textit{why} a computation fails. We introduce a white-box method: {\bf Circuit-based Reasoning Verification (CRV)}. We hypothesize that attribution graphs of correct CoT steps, viewed as \textit{execution traces} of the model's latent reasoning circuits, possess distinct structural fingerprints from those of incorrect steps. By training a classifier on structural features of these graphs, we show that these traces contain a powerful signal of reasoning errors. Our white-box approach yields novel scientific insights unattainable by other methods. (1) We demonstrate that structural signatures of error are highly predictive, establishing the viability of verifying reasoning directly via its computational graph. (2) We find these signatures to be highly domain-specific, revealing that failures in different reasoning tasks manifest as distinct computational patterns. (3) We provide evidence that these signatures are not merely correlational; by using our analysis to guide targeted interventions on individual transcoder features, we successfully correct the model's faulty reasoning. Our work shows that, by scrutinizing a model's computational process, we can move from simple error detection to a deeper, causal understanding of LLM reasoning.
}

\date{\today}
\correspondence{\email{zheng.zhao@ed.ac.uk}; \email{canceddan@acm.org}}

\metadata[Code]{\url{https://github.com/facebookresearch/CRV}}

\begin{document}

\maketitle

\section{Introduction}
\label{section:intro}

Chain-of-Thought (CoT;~\citealp{wei2022chain, Kojima2022LargeLM}) prompting has proven to be a powerful method for boosting the performance of Large Language Models (LLMs). This capability is now central to the latest generation of reasoning models, such as DeepSeek-R1 \citep{deepseekai2025deepseekr1} and OpenAI's o1 \citep{openai2024openaio1card}. Despite this success, a fundamental vulnerability persists across the spectrum of these systems: the reasoning process itself is sometimes flawed \citep{turpin-etal-2023-language, li2025thinkingfail, arcuschin2025chainofthought, lindsey2025biology, chen2025reasoningmodelsdontsay}.

This reliability gap has spurred research into automated verification. Current methods fall into two main categories. \textit{Black-box} approaches analyze the generated text or final logit distribution \citep{jacovi-etal-2024-chain, wang-etal-2025-chain, baker2025monitoringreasoningmodelsmisbehavior}. 
\textit{Gray-box} approaches look at the model's internal state, using simple probes on raw activations or analyzing the trajectory of hidden states \citep{xie2024calibrating, zhang2025reasoning, afzal-etal-2025-knowing, bi2025cot, wang2025latent}. While insightful, these methods are fundamentally limited; they can \textit{detect} that a model's internal state is correlated with an error, but not \textit{explain why} the underlying computation leads to an error.

This limitation motivates a deeper, more mechanistic approach. We postulate that models implement latent algorithms that solve specific tasks through specialized subgraphs, or {\bf circuits} \citep{olah2020zoom, elhage2021mathematical}. From this perspective, a reasoning failure is not merely an erroneous state, but a flaw in the execution of a latent algorithm. To diagnose such flaws requires inspecting the underlying computational process, akin to examining an \textit{execution trace} in classical software. We propose to approximate this trace by constructing an {\bf attribution graph}~\citep{dunefsky-etal-2024-transcoders}---a structural representation of the causal information flow between model components. 

For such a graph to serve as a meaningful trace, its components must be interpretable. We therefore first create an interpretable surrogate model by replacing its standard MLP modules with trained {\bf transcoders}~\citep{dunefsky-etal-2024-transcoders}. We then construct and analyze attribution graphs over the sparsely activating features of such surrogate model~\citep{ameisen2025circuit}. Finally, to formally test whether these traces contain a detectable signal of error, we train a diagnostic classifier on their structural properties. This entire methodology, which we call {\bf Circuit-based Reasoning Verification (CRV)}, is thus designed as a scientific instrument to investigate our central hypothesis: \textit{{that reasoning failures manifest as detectable structural signatures on their computational execution traces, which can be leveraged for automated verification.}}

As a scientific instrument, CRV requires a controlled experimental setting. While advanced reasoning models employ complex mechanisms like search and backtracking, their convoluted reasoning paths can obscure the fundamental computations of a single reasoning step. Our work therefore focuses on standard, instruction-tuned models generating autoregressive CoT, as this paradigm provides a clearer window into the primitive computations that underpin emergent reasoning. While our approach, despite being effective, is too computationally intensive to be intended as a practical, drop-in verifier, it yields novel scientific insights unattainable by other methods. Our main contributions are therefore not just about performance, but about understanding:
\begin{itemize}
    \item We introduce Circuit-based Reasoning Verification, a white-box method for analyzing reasoning failures, showing that verifying reasoning via its computational graph is feasible.
    \item We find that the structural signatures of error are highly {domain-specific}, revealing that failures in executing different reasoning tasks manifest as distinct computational patterns.
    \item We establish the causal role of these error signatures, successfully correcting faulty reasoning via targeted interventions on individual transcoder features.
    \item To support future research, we release datasets with step-level correctness labels for CoT reasoning on synthetic and real-world tasks, along with our trained transcoders.\footnote{\url{https://github.com/facebookresearch/CRV}}
\end{itemize}

\section{Problem Formulation and Preliminaries}

\subsection{Problem Statement}

Let an LLM generate a Chain-of-Thought $S = (s_1, s_2, \dots, s_m)$ to solve a problem, where each step $s_i$ is a sequence of tokens. During the generation of step $s_i$, the underlying model produces a computational state $\mathcal{M}_i$. From this state, we construct an {\bf attribution graph} $G_i = (\mathcal{V}, \mathcal{E})$, where vertices $\mathcal{V}$ represent interpretable features and tokens, and edges $\mathcal{E}$ represent the causal influence between them (see Section~\ref{sec:crv_method}). From each graph $G_i$, we extract a fixed-size feature vector $\mathbf{x}_i = \phi(G_i)$, where $\phi$ is a feature extraction function designed to capture the graph's structural properties. We term this vector the step's \textit{structural fingerprint}. Our goal is to learn a diagnostic classifier $f_\theta$ that takes this structural fingerprint as input to predict the correctness of the reasoning step:
$$ \hat{y}_i = f_\theta(\mathbf{x}_i) $$
where $\hat{y}_i \in \{\text{correct}, \text{incorrect}\}$.

\subsection{Preliminaries: Circuits in Transformers}

The term ``circuit'' in mechanistic interpretability refers to a specific subgraph within a neural network that implements a human-understandable algorithm~\citep{olah2020zoom}. In Transformers~\citep{vanswani-etal-2017-attention}, these circuits are composed of attention heads and MLP computations. 
Our work is conceptually motivated by the prospect of finding patterns distinguishing sound and faulty activations of circuits involved in reasoning. While our method does not observe these circuits directly, our hypothesis is that they cast detectable \textit{structural fingerprints} onto the attribution graphs we construct. A primary goal of our subsequent analysis is therefore to interpret the graph-based features that are most predictive of failure as the signatures of these underlying error patterns.

\subsection{Preliminaries: Transcoders for Interpretable Features}

A significant challenge in analyzing model activations is their high dimensionality and lack of direct interpretability. A powerful approach to this challenge is to learn a sparse, overcomplete basis for these activations using a sparse autoencoder (SAE;~\citealp{cunningham2023sparse}). An SAE is trained to reconstruct an activation vector $x \in \mathbb{R}^d$ from a much higher-dimensional, but mostly zero, feature vector $f \in \mathbb{R}^{D}$, where $D \gg d$. The elements of $f$ correspond to a set of learned, interpretable features, sparsely activated by inputs. While the canonical SAE objective is to reconstruct its own input ($f(x) \approx x$), our work leverages a variant known as a {transcoder} \citep{dunefsky-etal-2024-transcoders}, which is instead trained to approximate the \textit{input-output function} of a target component, such as an MLP ($f(x) \approx \text{MLP}(x)$). This approach makes the transcoder a true \textit{functional substitute} for the original module. Its objective is not mere reconstruction, but the emulation of a computational step in an interpretable, sparsely activated basis. By replacing a model's standard MLP module with a trained transcoder, we force its intermediate computations to be represented not by a dense vector, but by a sparse combination of these meaningful features. 

\begin{figure}[t]
\centering
\includegraphics[width=0.9\textwidth]{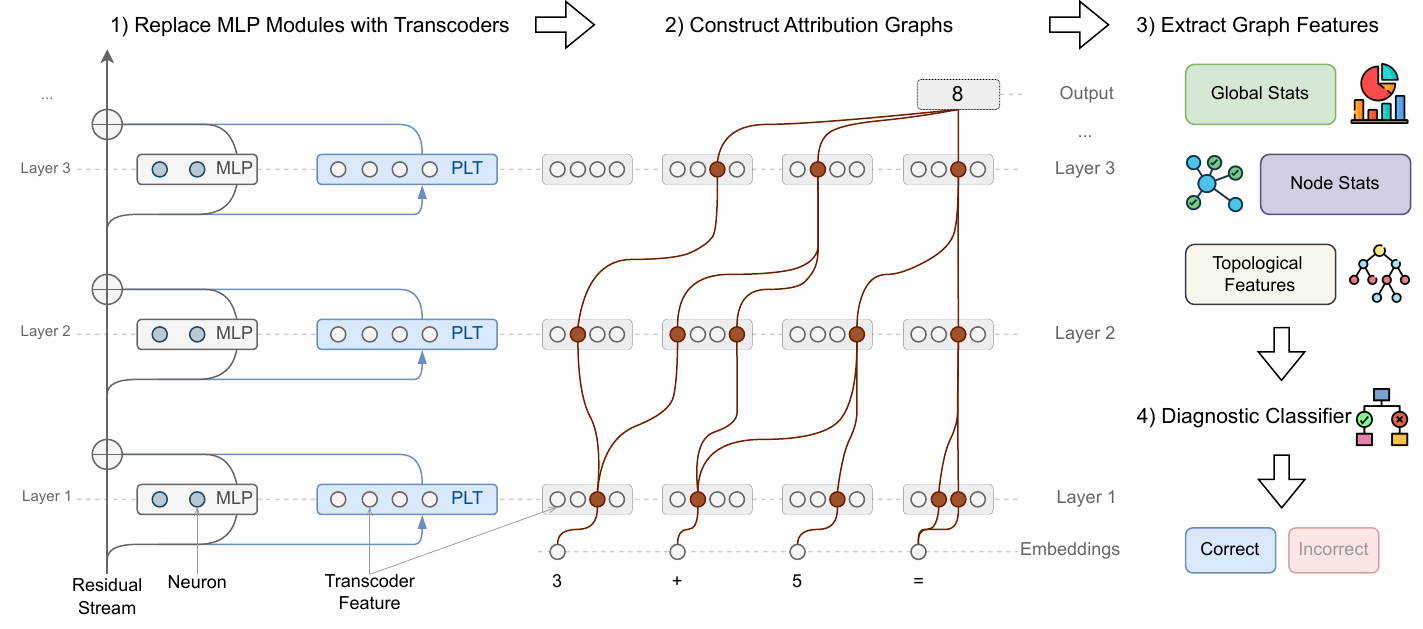} 
\caption{The CRV pipeline. (1) The LLM's MLP modules are replaced with per-layer transcoders (PLTs), making it interpretable. (2) For a given CoT step, we generate an attribution graph capturing causal flow between interpretable features and model components. (3) Structural features are extracted from this graph, and (4) fed to a diagnostic classifier to predict the step's correctness. }
\label{fig:overview}
\end{figure}

\section{Methodology}
\label{sec:method}

Unlike in Process Reward Modeling (PRM), where the goal is limited to judging the correctness of a reasoning step, we take the perspective of a model developer interested in debugging reasoning failures in a specific model to which they have full access. We introduce {\bf Circuit-based Reasoning Verification (CRV)}, a method for detecting flawed reasoning by analyzing its structural fingerprint. 

\subsection{Dataset Curation and Step-Level Annotation}
A prerequisite for developing our method is a dataset with reliable step-level correctness labels. Furthermore, our white-box methodology imposes a critical requirement that distinguishes our data needs from prior work. Since CRV analyzes the causal computational graph that produces a reasoning step, we must capture the full internal state of our specific model during the generation process. Consequently, existing text-only datasets such as PRM800K~\citep{lightman2024lets} and REVEAL~\citep{jacovi-etal-2024-chain}, which provide static `(text, label)' pairs and are designed for training black-box verifiers, are incompatible with our mechanistic approach. We must generate and label our own model's CoT outputs to create the necessary `(text, label, computational trace)' tuples for analysis. We therefore created a new benchmark covering both controlled synthetic tasks and the real-world GSM8K dataset~\citep{Cobbe2021TrainingVT}.

\paragraph{\bf Synthetic Datasets (Boolean and Arithmetic).} To study reasoning failures in a controlled environment, we generated two datasets. The first involves evaluating complex boolean expressions, while the second involves multi-step arithmetic problems. The motivation for these datasets is the unambiguous ground truth: the correctness of any step in the reasoning chain (e.g., ``15 + 7 = 22'') can be verified automatically by a simple parser and evaluator. This allows us to generate a large, labeled dataset for initial training and analysis. Furthermore, these tasks are intrinsically compositional, and the complexity of samples can be fully controlled. Further details are provided in Appendix~\ref{sec:appendix_dataset}.

\paragraph{\bf Step-Level Annotation for GSM8K.}
Annotating a real-world dataset like GSM8K is challenging. To scale, we used a semi-automated process with a stronger LLM (e.g., Llama 3.3 70B Instruct) as an expert judge. For each CoT, the judge evaluated step correctness given the full problem context. We validated these labels through manual review of a substantial subset, yielding a high-fidelity dataset for real-world reasoning. Further details are provided in Appendix~\ref{sec:appendix_dataset}.

\subsection{Circuit-based Reasoning Verification (CRV)}
\label{sec:crv_method}
CRV is a four-stage pipeline designed to classify the correctness of a CoT step by analyzing the computational graph of a modified, interpretable LLM. An overview is presented in Figure \ref{fig:overview}.

\subsubsection{Step 1: Replacing MLPs with Interpretable Transcoders}

The foundation of CRV is an architectural modification that makes the target LLM interpretable. For each MLP module in the model, we train a corresponding transcoder on a large, diverse dataset of activations harvested from the original LLM.\footnote{This is also referred as per-layer transcoders (PLTs) by \cite{ameisen2025circuit}.} The training objective combines an L2 reconstruction loss with a TopK activation function, which enforces sparsity by preserving only the k-largest feature activations. Once trained, we replace the MLP module for each layer in the LLM with its corresponding transcoder. The forward pass of the model is now forced to flow through these sparse, interpretable bottlenecks. All subsequent analysis is performed on this modified, interpretable replacement model. Full details of the transcoder architecture and training are provided in Appendix~\ref{sec:appendix_transcoder}.

\subsubsection{Step 2: Constructing Step-Level Attribution Graphs}

With our transcoder-infused replacement model, we require a principled method to trace information flow and construct a causal graph of the computation. To this end, we adapt the recent circuit analysis methodology of~\cite{dunefsky-etal-2024-transcoders}. Applying their greedy path-finding algorithm allows us to trace high-attribution connections backward from the final logits, yielding a sparse, weighted, directed graph $G_i = (\mathcal{V}, \mathcal{E})$ for each reasoning step $s_i$. This graph represents the core computational subgraph, where the nodes $\mathcal{V}$ are the disjoint union of input tokens, active transcoder features, and output logits. The directed edges $\mathcal{E}$ represent the high-attribution causal pathways between these components (e.g., from an early-layer feature to a later-layer feature, or from a feature to a logit), with weights quantifying the strength of their influence. For a complete derivation and description of the circuit-finding algorithm, we refer the reader to the original work~\citep{dunefsky-etal-2024-transcoders}.\footnote{We use implementation from \cite{circuit-tracer} for computing attribution graphs in our work.}

\subsubsection{Step 3: Extracting Interpretable Graph Features}

From each attribution graph $G_i$, we extract a fixed-size feature vector $\mathbf{x}_i$ as a structural fingerprint of the computation. We prune the graph to its most influential components, retaining nodes and edges accounting for a threshold (e.g., 80\%) of total influence to the final logits. The feature set, calculated on this pruned subgraph (unless stated otherwise), is organized into three hierarchical levels.

\paragraph{\bf Global Graph Statistics:} These features capture a high-level summary of the computational subgraph, including the count of active feature nodes after pruning and the final logit probability and entropy. They provide a coarse measure of the computation's complexity and uncertainty.

\paragraph{\bf Node Influence and Activation Statistics:} This group quantifies the properties of the interpretable feature nodes. We compute statistics (mean, max, std) on their activation values and influence scores. This helps distinguish computations driven by a few highly active, decisive features from those driven by a diffuse combination of many weak features. We also include a histogram of active features by layer, which characterizes the computational depth of the reasoning step.

\paragraph{\bf Topological and Path-Based Features:} To analyze the structure of the information flow, we compute a rich set of topological features on the pruned subgraph. These include graph density, centrality measures (degree, betweenness) to identify computational hubs, and connectivity metrics.

This comprehensive feature set provides the foundation for our diagnostic classifier. A full list and detailed motivation for each feature is provided in Appendix~\ref{app:attribution_feature}.

\subsubsection{Step 4: Diagnostic Classifier}
For the final classification step, we use a Gradient Boosting Classifier (GBC) trained on the extracted feature vectors: $f_\theta(\mathbf{x}_i) = \hat{y}_i$. GBC suits for our heterogeneous, tabular features and provides robust feature importance measures, which we leverage to identify the most predictive structural properties of error circuits. We also benchmark against several alternative classifiers in Appendix~\ref{app:appendix_classifiers}.

\section{Experiments}
\label{sec:experiments}

We conduct a series of experiments designed to validate the central hypothesis of our work: \emph{that the attribution graphs of reasoning steps contain a rich, structural signal of their correctness}. Our evaluation is structured around three primary research questions. First, we investigate whether CRV's white-box approach significantly outperforms a comprehensive suite of gray-box and black-box baselines in verification accuracy and test its robustness to domain shifts and increasing task difficulty {\bf (RQ1)}. Next, we analyze our trained models to identify which specific computational structures within the graph are most predictive of failure, moving from detection to mechanistic understanding {\bf (RQ2)}. 
Finally, we conduct exploratory studies to assess if these mechanistic insights can be used to perform targeted, causal interventions that correct faulty reasoning {\bf (RQ3)}.

\subsection{Experimental Setup}
\label{sec:experimental_setup}
\paragraph{\bf Models and Datasets.}
Our experiments are conducted on the {\bf Llama 3.1 8B Instruct} model \citep{llama3modelcard}. We select the instruction-tuned variant, as its prompt-following optimization is critical for reliably eliciting the CoT reasoning traces for our analysis. This model is then modified with our trained transcoders as described in Section~\ref{sec:method}. We evaluate performance on our three datasets: Synthetic (Boolean), Synthetic (Arithmetic), and the annotated GSM8K benchmark. 

\paragraph{\bf Baselines.}
We compare CRV against two categories of baselines. First, black-box methods that use the final logit distribution: Maximum Softmax Probability (MaxProb), Perplexity (PPL), Entropy, Temperature Scaling (Temp. Scaling;~\citealp{shih2023longhorizontemperaturescaling}), and Energy~\citep{liu-etal-2020-energy}. Second, gray-box methods that operate on internal states. This includes trajectory-based methods that analyze hidden state dynamics across layers, such as Chain-of-Embedding (with its real-space CoE-R and complex-space CoE-C variants;~\citealp{wang2025latent}) and CoT-Kinetics~\citep{bi2025cot}, as well as a standard logistic regression probe (LR Probe) and a two-layer multilayer perceptron probe (MLP Probe) trained on the step's average hidden state.\footnote{We also evaluated a last-token probe, but found that using the average representation yielded slightly better performance.} While CoE and CoT-Kinetics were originally designed for full CoT evaluation, they prove to be strong step-level baselines. All implementation details are deferred to Appendix~\ref{app:baseline_classifiers}.

\paragraph{\bf Evaluation Metrics.}
We use AUROC, FPR@95, and AUPR to evaluate verifier performance. As our goal is the detection of reasoning failures, we treat the incorrect label as the positive class for all metric calculations. AUROC assesses how well the method ranks correct versus incorrect steps across thresholds. AUPR captures the precision-recall trade-off for the positive (incorrect) class. FPR@95 measures the false positive rate when 95\% of positives are correctly identified, reflecting reliability under strict conditions; a lower score indicates the verifier can detect most errors with minimal false alarm. Together, these metrics provide complementary views of performance. Precise mathematical formulations and implementation details for all metrics are provided in Appendix~\ref{app:metric_definitions}.

\begin{table}[t]
\centering
\resizebox{\textwidth}{!}{%
\begin{tabular}{@{}ll ccc ccc ccc@{}}
\toprule
\multirow{2}{*}{{\bf Paradigm}} & \multirow{2}{*}{{\bf Method}} & \multicolumn{3}{c}{{\bf Synthetic (Boolean)}} & \multicolumn{3}{c}{{\bf Synthetic (Arithmetic)}} & \multicolumn{3}{c}{{\bf GSM8K}} \\
\cmidrule(l){3-5} \cmidrule(l){6-8} \cmidrule(l){9-11} 
 & & AUROC $\uparrow$ & AUPR $\uparrow$ & FPR@95 $\downarrow$ & AUROC $\uparrow$ & AUPR $\uparrow$ & FPR@95 $\downarrow$ & AUROC $\uparrow$ & AUPR $\uparrow$ & FPR@95 $\downarrow$ \\ \midrule
\multirow{4}{*}{\textit{Black-Box}} & MaxProb & \underline{58.81} & 0.34 & 95.20 & 61.87 & 1.81 & 84.98 & 54.91 & 7.99 & 91.86 \\
 & PPL & 57.37 & 0.29 & 91.02 & 60.19 & 1.68 & 85.52 & 55.46 & 8.12 & 90.69 \\
 & Entropy & 53.56 & 0.24 & 97.55 & 60.03 & 1.52 & 85.40 & 56.67 & 7.29 & 87.08 \\
 & Temp. Scaling & 58.77 & 0.36 & 91.41 & 59.67 & 1.66 & 86.96 & 54.42 & 8.24 & 92.28 \\
 & Energy & 51.08 & 0.28 & 95.11 & \underline{76.45} & \underline{5.59} & 73.86 & \underline{62.55} & 9.11 & \underline{86.34} \\\midrule
\multirow{2}{*}{\textit{Gray-Box}} & CoE-R & 53.17 & 0.33 & 92.85 & 58.47 & 1.93 & 76.68 & 52.38 & 8.34 & 96.20 \\
 & CoE-C & 51.03 & \underline{0.38} & 92.07 & 69.39 & 3.03 & \underline{63.33} & 53.57 & \underline{10.80} & 96.33 \\ 
 & CoT-Kinetics & 53.62 & 0.24 & 97.13 & 60.83 & 1.58 & 85.09 & 56.54 & 7.35 & 86.83 \\
 & LR Probe & 52.91 & 0.25 & \underline{88.42} & 54.22 & 1.50 & 91.90 & 55.86 & 7.99 & 90.32 \\ 
 & MLP Probe & 53.63 & 0.26 & 88.56 & 54.41 & 1.30 & 90.98 & 56.02 & 8.63 & 93.94 \\ \midrule
{\bf White-Box} & {\bf CRV (Ours)} & {\bf 75.87} & {\bf 0.97} & {\bf 79.17} & {\bf 92.47} & {\bf 28.92} & {\bf 37.09} & {\bf 70.17} & {\bf 14.3} & {\bf 79.61} \\ \bottomrule
\end{tabular}
}
\caption{Verification performance. Arrows indicate preferred direction ($\uparrow$ higher is better, $\downarrow$ lower is better). {\bf Best} and \underline{second-best} results are highlighted for each metric. 
The low AUPR on the Boolean dataset reflects extreme label imbalance, with the incorrect label only 0.2\% (Appendix~\ref{app:dataset_statistics}).
}
\label{tab:main_results}
\end{table}

\subsection{Verification Performance and Robustness (RQ1)}

We first address RQ1 by evaluating CRV against all baselines on the task of reasoning step verification and then probing its robustness under more challenging conditions. 

\paragraph{\bf Main Verification Performance.}
The results, presented in Table~\ref{tab:main_results}, provide strong empirical support for our central hypothesis: that the structural signatures present in a reasoning step's computational trace contain a directly verifiable signal of its correctness. CRV consistently outperforms all black-box and gray-box baselines across every dataset and metric. The strength of this structural signal is particularly evident on the synthetic datasets. On the Arithmetic task, for instance, CRV achieves an AUROC of 92.47, a significant leap over the strongest baseline score of 76.45. This advantage in reliability is further underscored by the FPR@95, where CRV reduces the false positive rate to 37.09\% from the baseline's 63.33\%. The performance gap is most pronounced on these structured, synthetic datasets. We hypothesize that the structured nature of algorithmic reasoning induces highly consistent execution traces for valid solutions. Consequently, the structural signatures of error manifest as more uniform deviations from this baseline, rendering them highly detectable.

\begin{wraptable}{r}{0.5\textwidth}
    \vspace{-12pt} 
    \centering
        
        \footnotesize 
        \setlength{\tabcolsep}{3pt} 
        \begin{tabular}{@{}ll ccc@{}}
        \toprule
        \multirow{2}{*}{{\bf Test Set}} & \multirow{2}{*}{{\bf Method (Train Set)}} & \multicolumn{3}{c}{{\bf Metrics}} \\
        \cmidrule(l){3-5} 
        & & AUROC $\uparrow$ & AUPR $\uparrow$ & FPR@95 $\downarrow$ \\ 
        \midrule
        \multirow{4}{*}{{Boolean}} & Baseline (MaxProb) & 58.81 & 0.34 & 95.20 \\
        & CRV (GSM8K) & 45.77 & 0.21 & 97.28 \\
        & CRV (Arithmetic) & {\bf 61.58}  & {\bf 0.51} & {\bf 87.55} \\
        \cmidrule(lr){2-5} 
        & CRV (Combined) & 65.18 & 0.65 & 80.15 \\
        & \textit{CRV (Boolean)} & \textit{75.87} & \textit{0.97} & \textit{79.17} \\
        \midrule
        \multirow{4}{*}{{Arithmetic}} & Baseline (Energy) & {\bf 76.45} & {\bf 5.59} & 73.86 \\
        & CRV (GSM8K) & 55.11 & 1.50 & 91.91 \\
        & CRV (Boolean) & 69.59 & 2.64 & {\bf 72.87} \\
        \cmidrule(lr){2-5}
        & CRV (Combined) & 90.51 & 30.57 & 44.23 \\
        & \textit{CRV (Arithmetic)} & \textit{92.47} & \textit{28.92} & \textit{37.09} \\
        \midrule
        \multirow{4}{*}{{GSM8K}} & Baseline (Energy) & {\bf 62.55} & {\bf 9.11} & {\bf 86.34} \\
        & CRV (Boolean) & 44.37 & 6.33 & 95.71 \\
        & CRV (Arithmetic) & 57.04 & 7.85 & 94.37 \\ 
        \cmidrule(lr){2-5} 
        & CRV (Combined) & 70.62 & 12.66 & 80.22 \\
        & \textit{CRV (GSM8K)} & \textit{70.17} & \textit{14.3} & \textit{79.61} \\
        \bottomrule
        \end{tabular}
        \caption{Cross-domain generalization performance. For each test dataset, we compare the strongest baseline (based on AUROC) against CRV trained \textit{in-domain} and out-of-domain. \textbf{Best} out-of-domain results are highlighted.}
        \label{tab:cross_domain}
\vspace{-14pt} 
\end{wraptable}

\paragraph{\bf Analysis of Cross-Domain Generalization.}
A key difference between CRV and most baselines is that its diagnostic classifier requires training. A critical question, therefore, is whether CRV learns domain-specific correlations or more fundamental, generalizable signatures of flawed reasoning. To test this, we conduct a comprehensive cross-domain evaluation. We train a CRV classifier on each of our three datasets individually and evaluate its zero-shot performance on the other two unseen tasks. We also evaluate a model trained on the union of all training sets to test if a single classifier can learn diverse error patterns simultaneously.

Table~\ref{tab:cross_domain} shows that CRV's zero-shot transfer is limited, suggesting that error fingerprints are highly task-specific. For example, CRV trained solely on Arithmetic achieves an AUROC of 57.04 on GSM8K, falling short of the Energy baseline. This reveals that errors in different reasoning tasks (e.g., boolean logic vs. natural language arithmetic) produce distinct structural patterns in the computational graph.

However, this domain specificity does not preclude the creation of a robust, multi-task verifier. As shown in the CRV (Combined) rows, when the classifier is exposed to diverse error signatures during training, performance rebounds. On the Arithmetic task, the combined model achieves an AUROC of 90.51, nearly matching the specialist in-domain model (92.47) and outperforming baselines. Remarkably, on GSM8K, the combined model (70.62 AUROC) slightly outperforms the in-domain specialist (70.17). This suggests that while the structural boundaries of error are distinct across tasks, they are not incompatible; a single classifier can learn the union of these complex ``failure geometries'', and in the case of GSM8K, potentially benefit from the inductive bias provided by synthetic logical tasks.

\begin{figure}[h]
    \centering
    \includegraphics[width=1.0\textwidth]{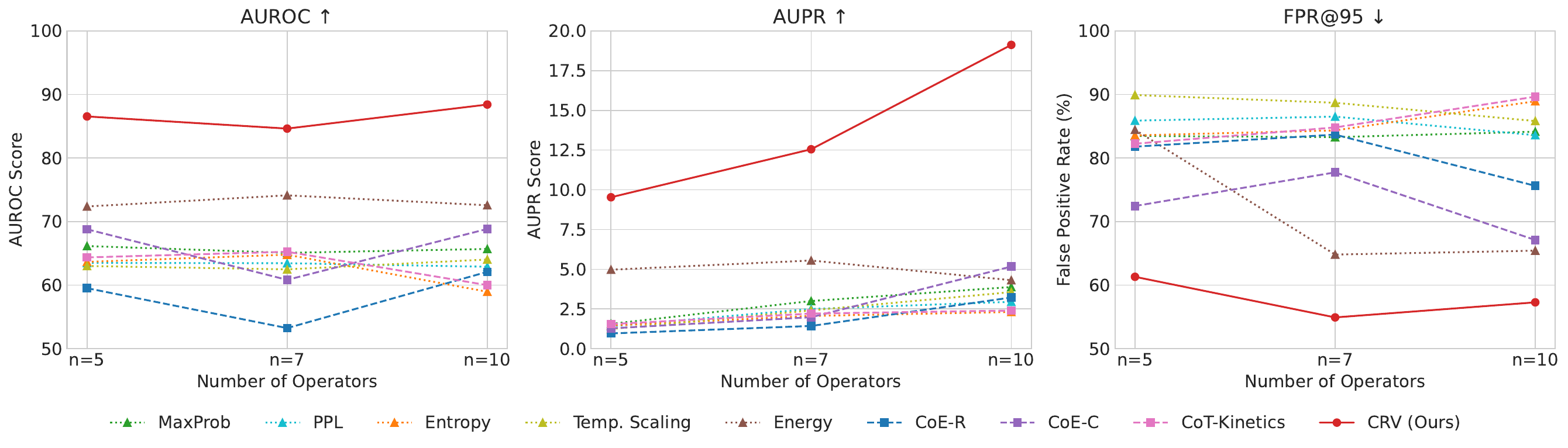}
    \caption{Performance of the step correctness predictors on the synthetic arithmetic task as a function of difficulty (number of operators). CRV retains a clear advantage as complexity increases.}
    \label{fig:difficulty_scaling}
\end{figure}

\paragraph{\bf Performance Under Increasing Difficulty.}
To further probe CRV's robustness, we analyze its performance on the synthetic arithmetic dataset as a function of problem complexity, controlled by the number of operators ($n \in \{5, 7, 10\}$).\footnote{We exclude $n=3$ as the model’s high accuracy yields too few incorrect examples for reliable evaluation.} Figure~\ref{fig:difficulty_scaling} plots the performance of CRV against key baselines across these difficulty levels. While most methods show stable AUROC and FPR@95, CRV maintains a consistent advantage across all difficulty levels. AUPR generally improves for all methods as difficulty rises because harder problems increase the proportion of incorrect examples (a condition to which AUPR is sensitive). Importantly, CRV’s advantage persists despite these shifts, highlighting the robustness of its structural signals across task difficulty and class balance.

\subsection{Mechanistic Analysis of Error Computations (RQ2)}

Having demonstrated CRV's predictive power, we now turn to its key advantage: interpretability. To address RQ2, we dissect our graph representation to identify structural ``fingerprints'' of error, from high-level feature ablation to fine-grained analysis of the most predictive structures.

\begin{wraptable}{r}{0.5\textwidth}
    \vspace{-12pt} 
    \centering
    \footnotesize
    \begin{tabular}{@{}l ccc@{}}
    \toprule
    \multirow{2}{*}{{\bf Feature Set}} & \multicolumn{3}{c}{{\bf Arithmetic}} \\
    \cmidrule(l){2-4} 
     & AUROC ↑ & AUPR ↑ & FPR@95 ↓ \\ 
    \midrule
    {\bf CRV (All three families)} & {\bf 92.47} & {\bf 28.92} & {\bf 37.09} \\ 
    \midrule
    \textit{Ablation:} \\
    \quad -- w/o Global Stats & 89.62 & 24.35 & 44.54  \\
    \quad -- w/o Node Stats & 88.31 & 23.25 & 49.07 \\
    \quad -- w/o Topological Stats & 90.89  & 26.83 & 39.19 \\ 
    \bottomrule
    \end{tabular}
    \caption{Leave-one-out ablation study on the Synthetic (Arithmetic) dataset.
    }
    \label{tab:ablation}
\vspace{-10pt} 
\end{wraptable}

\paragraph{\bf Ablation of Feature Families.}
A leave-one-out ablation study on the Synthetic (Arithmetic) dataset reveals a clear hierarchy of feature importance, as summarized in Table~\ref{tab:ablation}. The \textit{Node Influence \& Activation} features are demonstrably the most critical; their removal causes the most performance degradation across all metrics, most notably increasing FPR@95 by over 12 points. The \textit{Global Graph Statistics} also provide a substantial contribution. Interestingly, the \textit{Topological \& Path-Based} features appear least critical for this specific task, suggesting that the state of key local features is a more dominant signal than the holistic graph structure. Nevertheless, the full CRV model, which integrates all three signal types, is required to achieve optimal verification performance.

\begin{figure}[t!]
    \centering
    \begin{subfigure}[b]{0.19\textwidth}
        \centering
        \includegraphics[width=\linewidth]{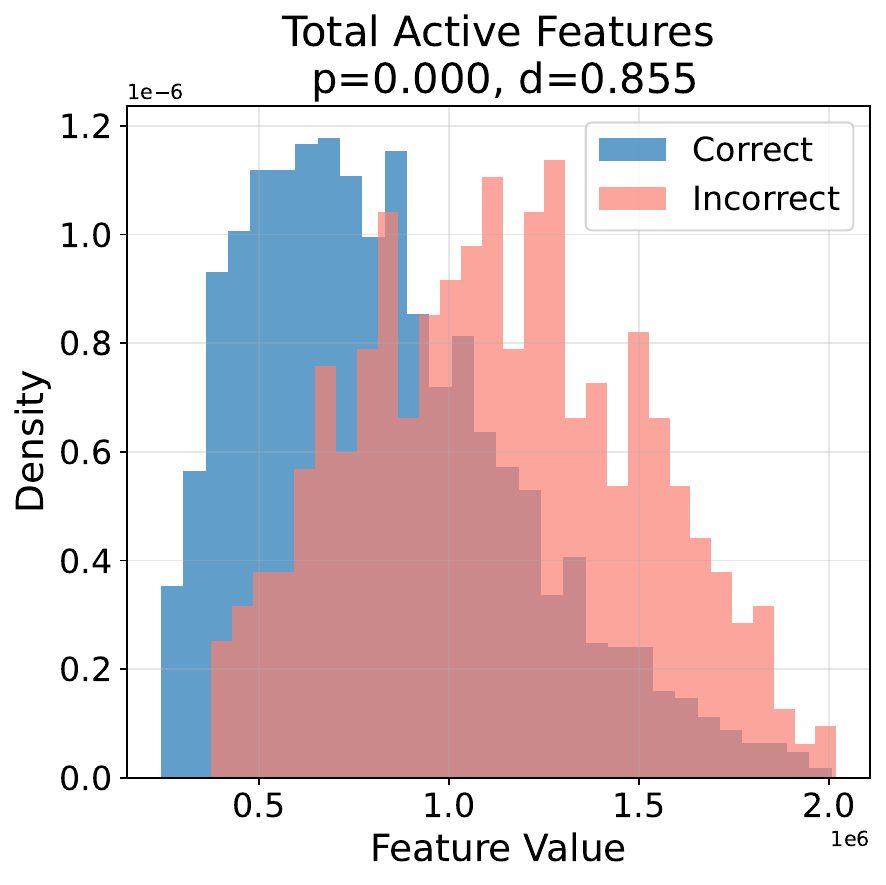} 
        \label{fig:dist_density}
    \end{subfigure}
    \hfill
    \begin{subfigure}[b]{0.19\textwidth}
        \centering
        \includegraphics[width=\linewidth]{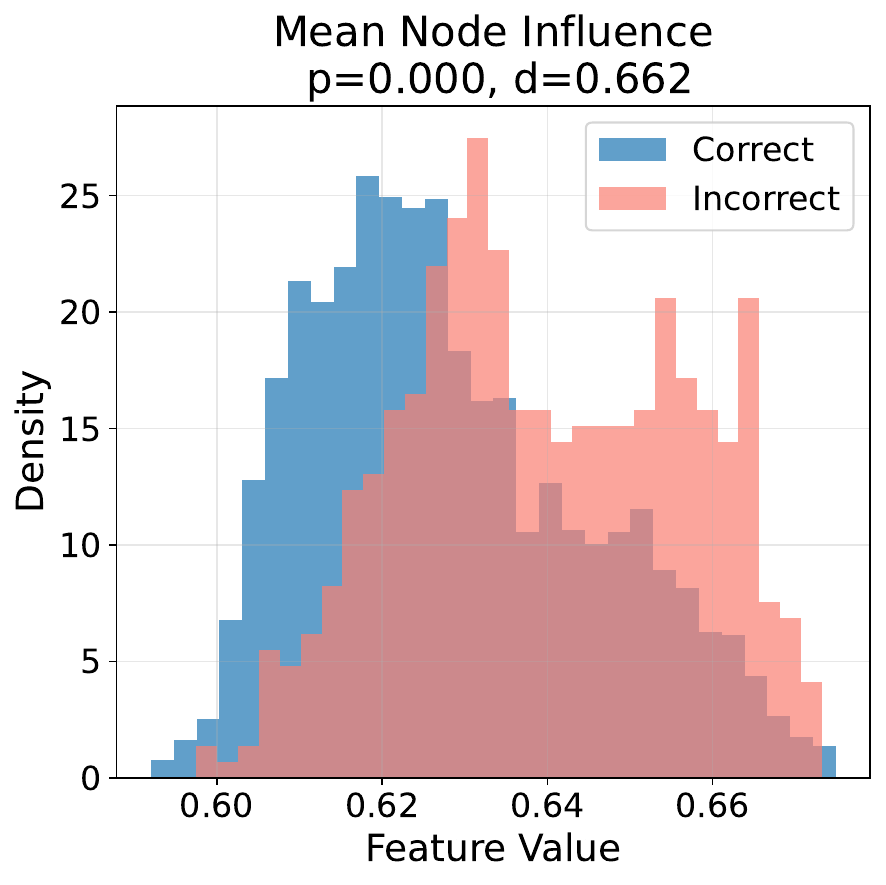} 
        \label{fig:dist_pathlength}
    \end{subfigure}
    \hfill
    \begin{subfigure}[b]{0.19\textwidth}
        \centering
        \includegraphics[width=\linewidth]{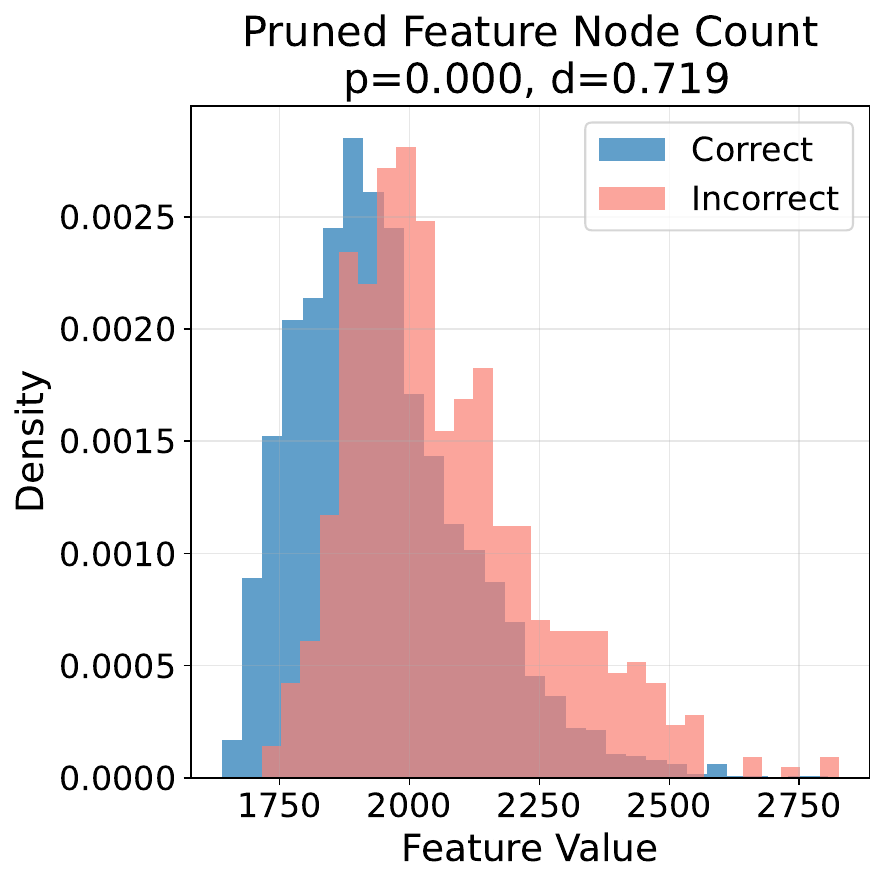} 
        \label{fig:dist_centrality}
    \end{subfigure}
    \hfill
    \begin{subfigure}[b]{0.19\textwidth}
        \centering
        \includegraphics[width=\linewidth]{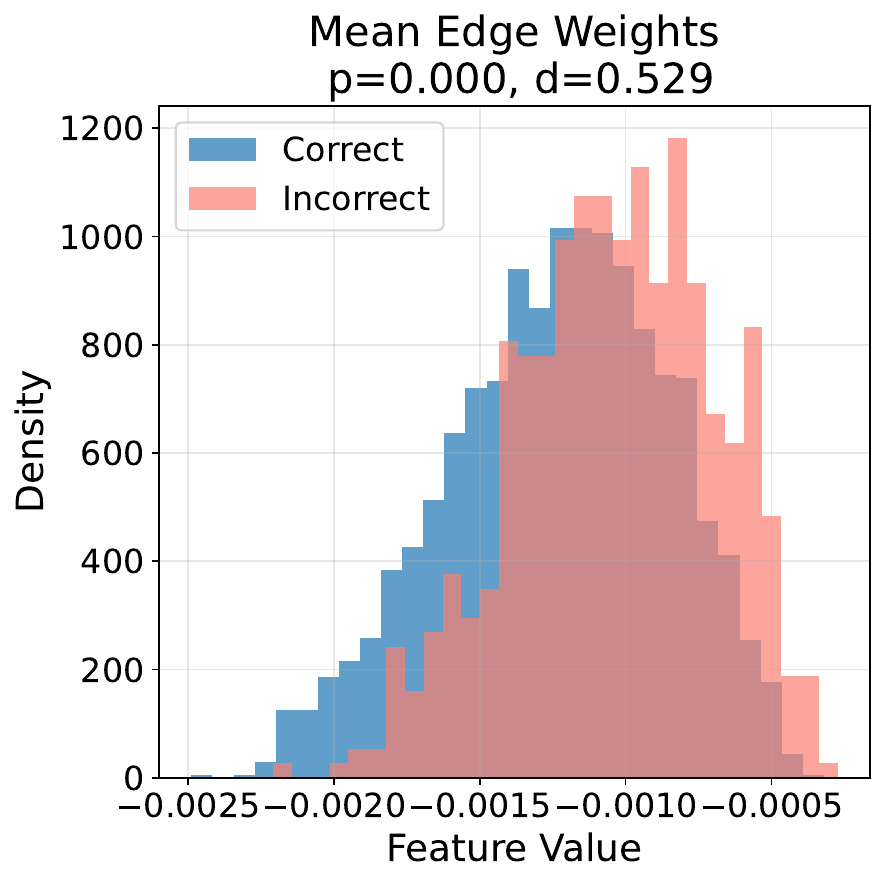} 
        \label{fig:dist_feature_count}
    \end{subfigure}
    \hfill
    \begin{subfigure}[b]{0.185\textwidth}
        \centering
        \includegraphics[width=\linewidth]{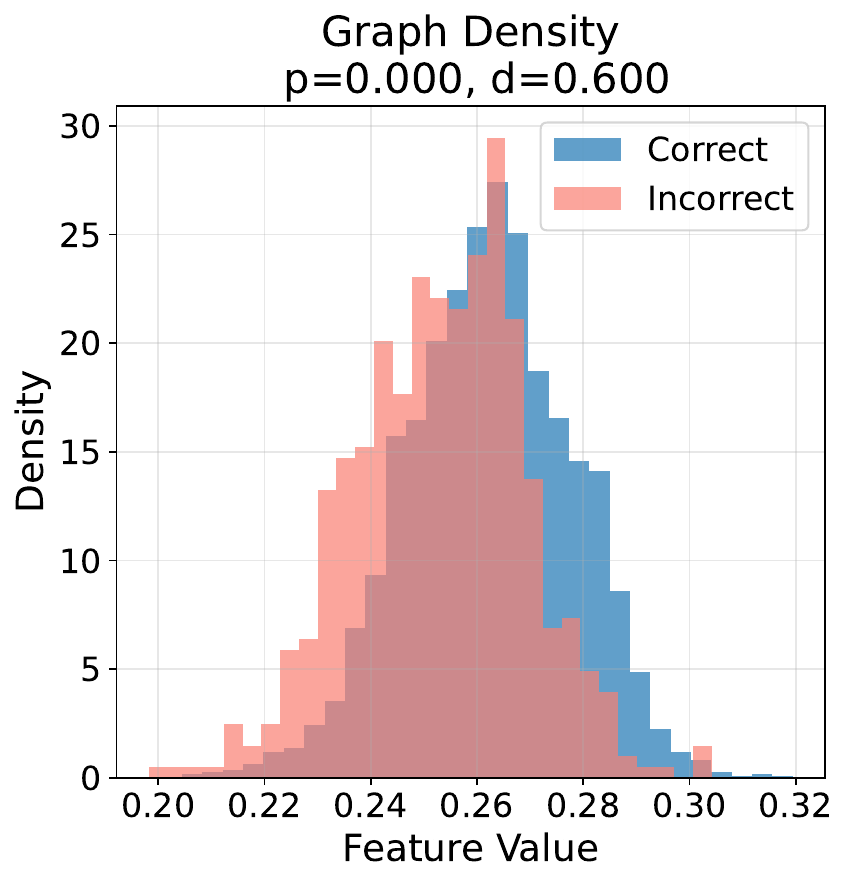} 
        \label{fig:dist_entropy}
    \end{subfigure}
    
    \caption{
        {\bf Topological Fingerprints of Error on GSM8K.} 
        Distributions of five selected graph features for correct (blue) vs. incorrect (red) reasoning steps. 
        The visual separation is statistically significant for each feature shown (independent t-test, $p < 0.001$) and represents a medium-to-large effect size (Cohen's d). This provides quantitative evidence that attribution graphs contain a clear, separable structural signal of a computation's integrity.
    }
    \label{fig:feature_distributions}
\end{figure}

\paragraph{\bf Visualizing the Structural Signatures of Error.}
To provide qualitative evidence for our hypothesis, we visualize the ``structural fingerprints'' learned by our classifier. Figure~\ref{fig:feature_distributions} shows distributions of five highly predictive features for correct versus incorrect GSM8K reasoning steps. Across diverse feature types, from graph topology (e.g., \textit{Graph Density}) to node statistics (e.g., \textit{Total Active Features}), distributions are clearly distinct. Similar patterns are observed on our synthetic datasets (see included in Appendix~\ref{app:additional_results}), confirming that the graph representation captures separable structural differences between valid and flawed computations.

\begin{wrapfigure}{r}{0.65\textwidth}
    \centering
    \begin{subfigure}[b]{0.32\linewidth}
        \centering
        \includegraphics[width=\linewidth]{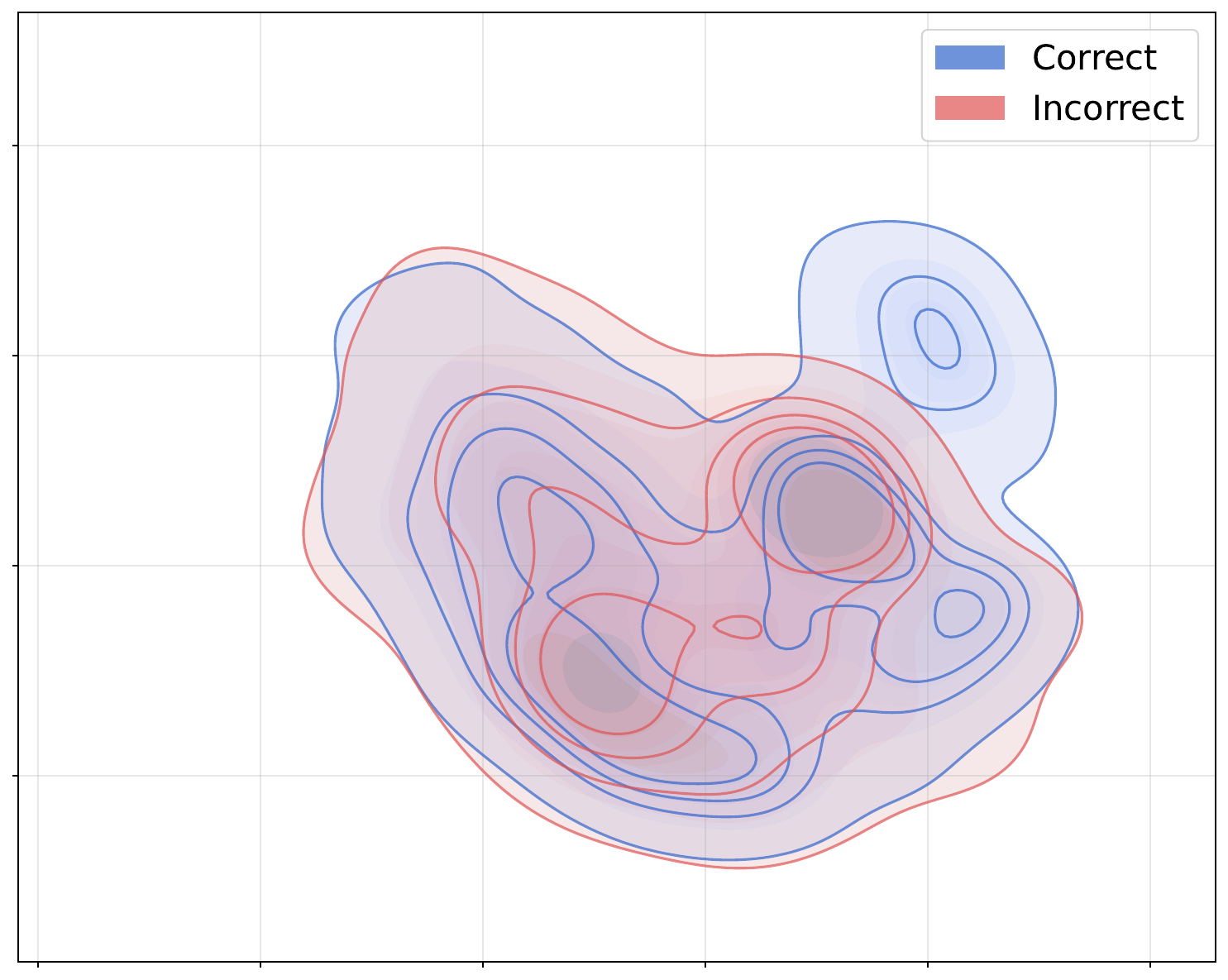} 
        \caption{Boolean}
        \label{fig:dist_bool}
    \end{subfigure}
    \hfill
    \begin{subfigure}[b]{0.32\linewidth}
        \centering
        \includegraphics[width=\linewidth]{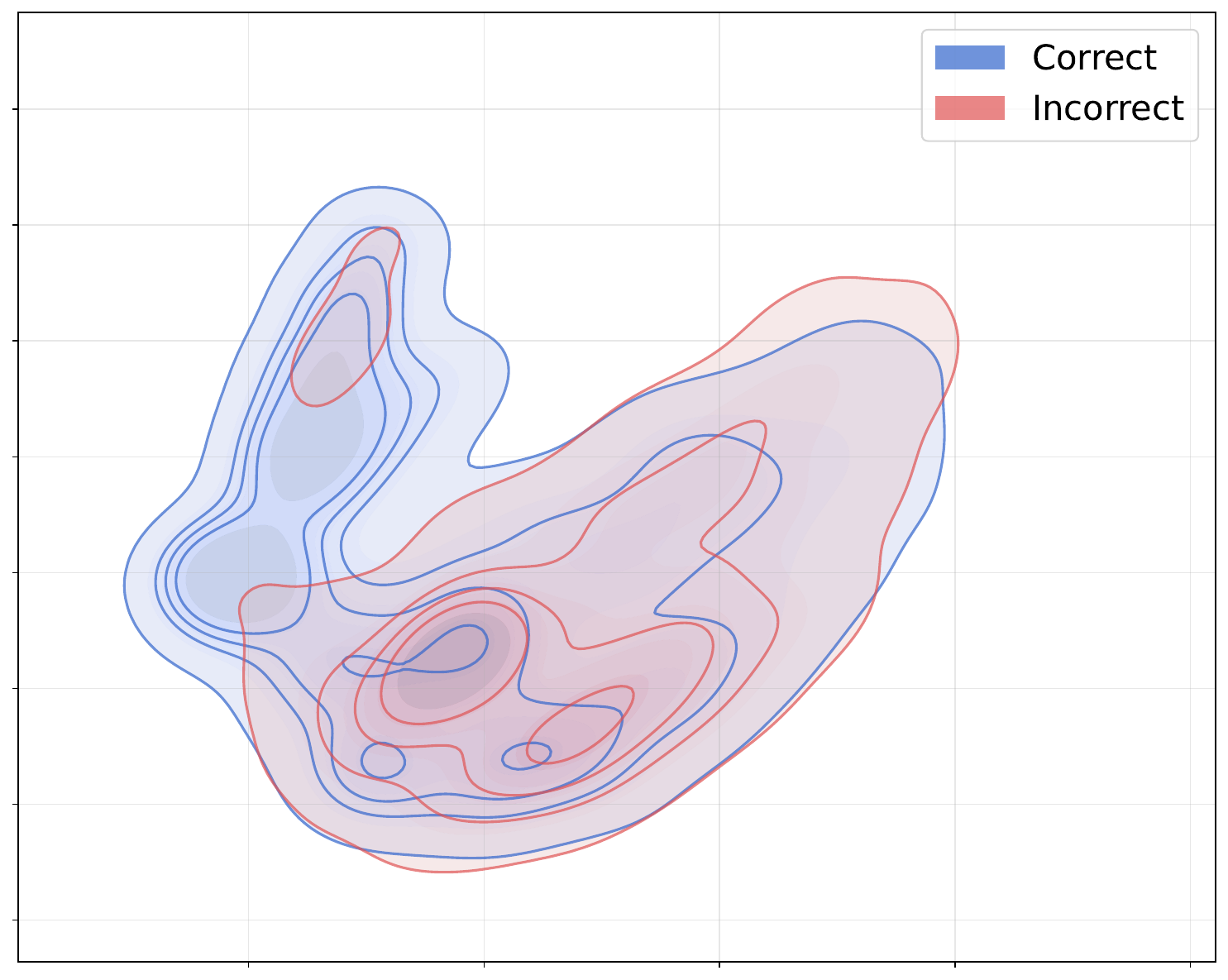} 
        \caption{Arithmetic}
        \label{fig:dist_arith}
    \end{subfigure}
    \hfill
    \begin{subfigure}[b]{0.32\linewidth}
        \centering
        \includegraphics[width=\linewidth]{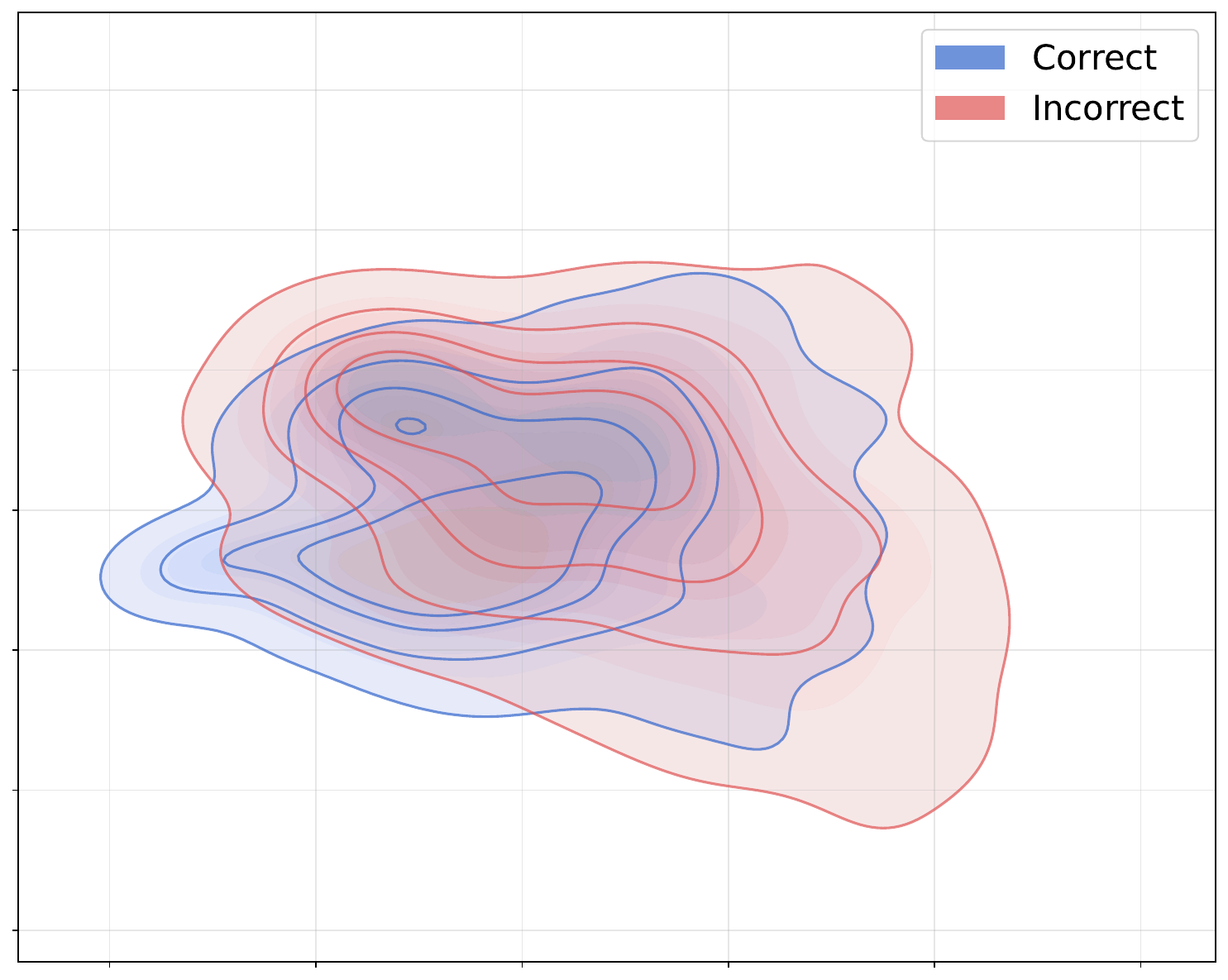} 
        \caption{GSM8K}
        \label{fig:dist_gsm8k}
    \end{subfigure}
    \caption{
        Distributions of features after PCA for correct (blue) vs. incorrect (red) reasoning steps. 
    }
    \label{fig:dist_comparison}
    \vspace{-10pt} 
\end{wrapfigure}

While individual features are predictive, CRV’s strength lies in their combination. To illustrate this, we project the full high-dimensional feature vectors into two dimensions via Principal Component Analysis (PCA). Figure~\ref{fig:dist_comparison} reveals that incorrect steps form a dense subset \textit{within} the broader distribution of correct steps.  

Crucially, correct steps also occupy a distinct region not shared by incorrect computations. This suggests many reasoning failures are computational ``near misses'', structurally similar to valid steps, which explains the overlap. Yet there exists a \textit{zone of computational integrity}, a region defined by structural properties accessible only to correct reasoning. A complementary t-SNE visualization in Appendix~\ref{app:additional_results} shows the same structure, confirming that CRV succeeds by learning the complex boundary of this zone rather than separating two simple clusters.

\subsection{From Diagnosis to Causal Intervention (RQ3)}

Finally, we conduct an exploratory study to test whether CRV's diagnostic insights can guide prescriptive interventions. A key advantage of our white-box approach is that failure-predictive graph features can be traced to specific components of the model’s computation. We illustrate this with a case study on the arithmetic task, correcting an order-of-operations error.

Given the expression \texttt{(7*((5+9)+7))}, the model incorrectly computed \texttt{7*14=98} in its second step (Table~\ref{tab:intervention_example}). CRV flagged this step as incorrect, and feature importance highlighted an unusually high activation of a late-layer transcoder feature. Tracing the signal back to the graph, we found a single highly active last-layer transcoder feature (ID 91814) strongly associated with multiplication.

We hypothesized that this premature activation caused the error, and performed a direct intervention. We re-ran the generation up to the point of failure and used a forward hook to manually clamp the activation of this specific \emph{multiplication} feature to zero. The effect was immediate: as shown in the right column of Table~\ref{tab:intervention_example}, with the premature multiply impulse suppressed, the model correctly generated the next step \texttt{14+7=21} and proceeded to the correct final answer.

\begin{table}[h]
\centering
\begin{tabular}{p{0.47\textwidth} p{0.47\textwidth}}
\toprule
{\bf Before Intervention (Incorrect)} & {\bf After Intervention (Correct)} \\
\midrule

\parbox[t]{\linewidth}{
    Evaluate the arithmetic expression below. \par
    ( 7 * ( ( 5 + 9 ) + 7 ) ) \par
    To evaluate this expression, we need to follow the order of operations (PEMDAS): \par
    1. Evaluate the expression inside the innermost parentheses: 5 + 9 = 14 \par
    2. \sethlcolor{myred}\hl{Multiply 7 by the result: 7 * 14 = 98} \par
    \sethlcolor{myred}\hl{3. Add 7 to the result: 98 + 7 = 105} \par
    \sethlcolor{myred}\hl{Therefore, the expression evaluates to 105.}
} &

\parbox[t]{\linewidth}{
    Evaluate the arithmetic expression below. \par
    ( 7 * ( ( 5 + 9 ) + 7 ) ) \par
    To evaluate this expression, we need to follow the order of operations (PEMDAS): \par
    1. Evaluate the expression inside the innermost parentheses: 5 + 9 = 14 \par
    2. \sethlcolor{myblue}\hl{Add 7 to the result: 14 + 7 = 21} \par
    \sethlcolor{myblue}\hl{3. Multiply 7 by the result 7 * 21 = 147} \par
    \sethlcolor{myblue}\hl{Therefore, the value of the expression is 147.}
} \\
\bottomrule
\end{tabular}
\caption{Side-by-side comparison of a reasoning trace \sethlcolor{myred}\hl{before} and \sethlcolor{myblue}\hl{after} causal intervention. The highlight indicates the point of divergence where suppressing a single \emph{multiplication} transcoder feature corrects the model's computational path.}
\vspace{-1.5ex}
\label{tab:intervention_example}
\end{table}

While not a general-purpose solution, this proof-of-concept is a crucial step. Beyond suppressing faulty features, we also corrected an error by \textit{amplifying} an under-active feature (Appendix~\ref{app:additional_results}). The success of both interventions provides closed-loop evidence that CRV’s structural signatures are causally implicated in errors, opening a promising direction for targeted model interventions. Crucially, because these interventions rely on the semantic interpretation of specific transcoder features (e.g., identifying a missing ``multiplication'' concept), they demonstrate the potential of CRV to serve as a foundation for interpretable debugging, moving beyond simple error flagging to actionable model repair.

\section{Related Work}

\paragraph{\bf CoT Prompting for Improved Reasoning.}
Chain-of-Thought (CoT) prompting was introduced by \cite{wei2022chain} to elicit more complex reasoning from LLMs through few-shot examples that demonstrate step-by-step problem solving. Subsequent work has expanded this idea in several directions. \cite{Kojima2022LargeLM} showed that even a simple zero-shot instruction such as ``Let’s think step by step'' can trigger coherent reasoning traces. While this reduces the need for handcrafted prompts, providing structured examples often remains beneficial. To scale this process, recent studies generate CoT exemplars synthetically~\citep{zhang2022automatic,shao2023synthetic,li2025structured}. Other work leverages test-time compute scaling to extend reasoning chains, enabling longer and more elaborate solutions~\citep{snell2024scaling}. For comprehensive surveys of CoT techniques and their applications, see~\cite{chu2023navigate} and ~\cite{chen2025towards}.

\paragraph{\bf Verifying and Improving CoT Reasoning.}
The transparency of CoT has also made it a focal point for research into model interpretability and reliability. While some work assumes reasoning traces are to some extent faithful representations of the model's internal process \citep{yeo2024interpretable, korbak2025chain}, a significant body of evidence highlights their unreliability \citep{arcuschin2025chainofthought, bentham2024chain, chen2025reasoningmodelsdontsay, turpin-etal-2023-language}. This has spurred a field of research dedicated to verifying and improving CoT traces. This research broadly investigates (i) the model's intrinsic ability to self-evaluate its reasoning steps \citep{zhang2025reasoning}, (ii) how to measure the faithfulness of a reasoning chain to the final answer \citep{lanham2023measuring,bi2025cot,tutek2025measuring}, and (iii) when reasoning steps are needed or useful \citep{bogdan2025thought,wang-etal-2025-chain}. A parallel line of work aims to improve reasoning chains through various forms of neuro-symbolic reasoning~\citep{lyu2023faithful}, correction~\citep{tyen2024llms}, uncertainty calibration~\citep{ji2025calibrating}, or by enforcing internal consistency~\citep{xie2024calibrating, wang2025latent}. A distinct approach involves training auxiliary models, such as Process Reward Models (PRMs), to assess step-level correctness and guide post-training \citep{lightman2024lets, wang-etal-2024-math, guan2025rstarmath}. While all these methods aim to improve reasoning outcomes, they primarily operate on the textual or hidden state representations. We are not aware of previous attempts to verify reasoning by analyzing the structural properties of its underlying computational graph.

\paragraph{\bf Mechanistic Interpretability of CoT Reasoning.}
Our work is most directly situated within the field of mechanistic interpretability, which seeks to reverse-engineer the algorithms learned by neural networks, moving beyond the surface-level analysis of CoT traces \citep{yeo2024interpretable, korbak2025chain, baker2025monitoringreasoningmodelsmisbehavior}. A central tenet of this field is that models develop specialized subgraphs, or circuits, to perform specific computations \citep{olah2020zoom}. Recent work has begun to apply this lens to reasoning, providing evidence that CoT traces are driven by latent learned algorithms \citep{venhoff2025basemodelsknowreason} and even characterizing specific self-verification circuits in toy models \citep{lee2025geometry}. Other research aims to improve performance by eliciting or steering behavioral circuits \citep{zhao2024steering, ward2025reasoning}. A particularly powerful and increasingly popular tool in this area is the use of sparse autoencoders (SAEs), which learn to decompose a model's dense activation vectors into a sparse basis of interpretable features \citep{bricken2023monosemanticity, cunningham2023sparse}. Our work builds directly on a variant, the transcoder \citep{dunefsky-etal-2024-transcoders}, which acts as a functional, interpretable substitute for an MLP module. While prior work has used transcoder-based attribution graphs to qualitatively analyze the faithfulness of CoT reasoning \citep{ameisen2025circuit}, our work is the first to operationalize this approach for automated verification. We move beyond visual inspection by systematically extracting quantitative, structural features from these graphs and demonstrating that they can be used to diagnose computational failures.

\section{Conclusion}

In this work, we introduced CRV, a white-box methodology for studying the computational structure of reasoning failures. By treating attribution graphs as execution traces of latent circuits, we showed that correct and incorrect reasoning leave distinct structural fingerprints. CRV revealed that these error signatures not only enable accurate verification but are also domain-specific, with failures in different reasoning tasks manifesting as distinct patterns. Moreover, targeted interventions on transcoder features demonstrated that these signatures are causally implicated, allowing us to correct faulty reasoning. Together, these findings establish CRV as a proof-of-concept for mechanistic analysis, showing that shifting from opaque activations to interpretable computational structure enables a causal understanding of how and why LLMs fail to reason correctly.

\section*{Acknowledgments}
We are grateful to Edan Toledo and Karen Hambardzumyan for their constructive discussions and insightful feedback on this project. We also thank Shuangrui Ding and Yunzhen Feng for their helpful input, and Megan Ung for assistance with setting up the computing environment.

\bibliographystyle{assets/plainnat}
\bibliography{paper}

@misc{Cobbe2021TrainingVT,
      title={Training Verifiers to Solve Math Word Problems}, 
      author={Karl Cobbe and Vineet Kosaraju and Mohammad Bavarian and Mark Chen and Heewoo Jun and Lukasz Kaiser and Matthias Plappert and Jerry Tworek and Jacob Hilton and Reiichiro Nakano and Christopher Hesse and John Schulman},
      year={2021},
      eprint={2110.14168},
      archivePrefix={arXiv},
      primaryClass={cs.LG},
      url={https://arxiv.org/abs/2110.14168}, 
}

@inproceedings{Kojima2022LargeLM,
author = {Kojima, Takeshi and Gu, Shixiang Shane and Reid, Machel and Matsuo, Yutaka and Iwasawa, Yusuke},
title = {Large language models are zero-shot reasoners},
year = {2022},
isbn = {9781713871088},
publisher = {Curran Associates Inc.},
address = {Red Hook, NY, USA},
booktitle = {Proceedings of the 36th International Conference on Neural Information Processing Systems},
articleno = {1613},
numpages = {15},
location = {New Orleans, LA, USA},
series = {NIPS '22}
}

@inproceedings{chu2023navigate,
    title = "Navigate through Enigmatic Labyrinth A Survey of Chain of Thought Reasoning: Advances, Frontiers and Future",
    author = "Chu, Zheng  and
      Chen, Jingchang  and
      Chen, Qianglong  and
      Yu, Weijiang  and
      He, Tao  and
      Wang, Haotian  and
      Peng, Weihua  and
      Liu, Ming  and
      Qin, Bing  and
      Liu, Ting",
    editor = "Ku, Lun-Wei  and
      Martins, Andre  and
      Srikumar, Vivek",
    booktitle = "Proceedings of the 62nd Annual Meeting of the Association for Computational Linguistics (Volume 1: Long Papers)",
    month = aug,
    year = "2024",
    address = "Bangkok, Thailand",
    publisher = "Association for Computational Linguistics",
    url = "https://aclanthology.org/2024.acl-long.65/",
    doi = "10.18653/v1/2024.acl-long.65",
    pages = "1173--1203"
}

@article{bricken2023monosemanticity,
       title={Towards Monosemanticity: Decomposing Language Models With Dictionary Learning},
       author={Bricken, Trenton and Templeton, Adly and Batson, Joshua and Chen, Brian and Jermyn, Adam and Conerly, Tom and Turner, Nick and others},
       year={2023},
       journal={Transformer Circuits Thread},
       note={https://transformer-circuits.pub/2023/monosemantic-features/index.html}
    }

@article{ameisen2025circuit,
  author={Ameisen, Emmanuel and Lindsey, Jack and Pearce, Adam and Gurnee, Wes and Turner, Nicholas L. and Chen, Brian and Citro, Craig and Abrahams, David and Carter, Shan and Hosmer, Basil and Marcus, Jonathan and Sklar, Michael and Templeton, Adly and Bricken, Trenton and McDougall, Callum and Cunningham, Hoagy and Henighan, Thomas and Jermyn, Adam and Jones, Andy and Persic, Andrew and Qi, Zhenyi and Ben Thompson, T. and Zimmerman, Sam and Rivoire, Kelley and Conerly, Thomas and Olah, Chris and Batson, Joshua},
  title={Circuit Tracing: Revealing Computational Graphs in Language Models},
  journal={Transformer Circuits Thread},
  year={2025},
  url={https://transformer-circuits.pub/2025/attribution-graphs/methods.html}
}

@inproceedings{yeo2024interpretable,
    title = "How Interpretable are Reasoning Explanations from Prompting Large Language Models?",
    author = "Wei Jie, Yeo  and
      Satapathy, Ranjan  and
      Goh, Rick  and
      Cambria, Erik",
    editor = "Duh, Kevin  and
      Gomez, Helena  and
      Bethard, Steven",
    booktitle = "Findings of the Association for Computational Linguistics: NAACL 2024",
    month = jun,
    year = "2024",
    address = "Mexico City, Mexico",
    publisher = "Association for Computational Linguistics",
    url = "https://aclanthology.org/2024.findings-naacl.138/",
    doi = "10.18653/v1/2024.findings-naacl.138",
    pages = "2148--2164"
}

@inproceedings{shao2023synthetic,
author = {Shao, Zhihong and Gong, Yeyun and Shen, Yelong and Huang, Minlie and Duan, Nan and Chen, Weizhu},
title = {Synthetic prompting: generating chain-of-thought demonstrations for large language models},
year = {2023},
publisher = {JMLR.org},
booktitle = {Proceedings of the 40th International Conference on Machine Learning},
articleno = {1273},
numpages = {70},
location = {Honolulu, Hawaii, USA},
series = {ICML'23}
}

@inproceedings{
zhang2022automatic,
title={Automatic Chain of Thought Prompting in Large Language Models},
author={Zhuosheng Zhang and Aston Zhang and Mu Li and Alex Smola},
booktitle={The Eleventh International Conference on Learning Representations},
year={2023},
url={https://openreview.net/forum?id=5NTt8GFjUHkr}
}

@article{li2025structured,
  title={Structured chain-of-thought prompting for code generation},
  author={Li, Jia and Li, Ge and Li, Yongmin and Jin, Zhi},
  journal={ACM Transactions on Software Engineering and Methodology},
  volume={34},
  number={2},
  pages={1--23},
  year={2025},
  publisher={ACM New York, NY}
}

@misc{chen2025towards,
      title={Towards Reasoning Era: A Survey of Long Chain-of-Thought for Reasoning Large Language Models}, 
      author={Qiguang Chen and Libo Qin and Jinhao Liu and Dengyun Peng and Jiannan Guan and Peng Wang and Mengkang Hu and Yuhang Zhou and Te Gao and Wanxiang Che},
      year={2025},
      eprint={2503.09567},
      archivePrefix={arXiv},
      primaryClass={cs.AI},
      url={https://arxiv.org/abs/2503.09567}, 
}

@misc{cunningham2023sparse,
      title={Sparse Autoencoders Find Highly Interpretable Features in Language Models}, 
      author={Hoagy Cunningham and Aidan Ewart and Logan Riggs and Robert Huben and Lee Sharkey},
      year={2023},
      eprint={2309.08600},
      archivePrefix={arXiv},
      primaryClass={cs.LG},
      url={https://arxiv.org/abs/2309.08600}, 
}

@article{olah2020zoom,
  author = {Olah, Chris and Cammarata, Nick and Schubert, Ludwig and Goh, Gabriel and Petrov, Michael and Carter, Shan},
  title = {Zoom In: An Introduction to Circuits},
  journal = {Distill},
  year = {2020},
  note = {https://distill.pub/2020/circuits/zoom-in},
  doi = {10.23915/distill.00024.001}
}

@inproceedings{wei2022chain,
author = {Wei, Jason and Wang, Xuezhi and Schuurmans, Dale and Bosma, Maarten and Ichter, Brian and Xia, Fei and Chi, Ed H. and Le, Quoc V. and Zhou, Denny},
title = {Chain-of-thought prompting elicits reasoning in large language models},
year = {2022},
isbn = {9781713871088},
publisher = {Curran Associates Inc.},
address = {Red Hook, NY, USA},
booktitle = {Proceedings of the 36th International Conference on Neural Information Processing Systems},
articleno = {1800},
numpages = {14},
location = {New Orleans, LA, USA},
series = {NIPS '22}
}

@inproceedings{
wang2025latent,
title={Latent Space Chain-of-Embedding Enables Output-free {LLM} Self-Evaluation},
author={Yiming Wang and Pei Zhang and Baosong Yang and Derek F. Wong and Rui Wang},
booktitle={The Thirteenth International Conference on Learning Representations},
year={2025},
url={https://openreview.net/forum?id=jxo70B9fQo}
}

@misc{bi2025cot,
      title={CoT-Kinetics: A Theoretical Modeling Assessing LRM Reasoning Process}, 
      author={Jinhe Bi and Danqi Yan and Yifan Wang and Wenke Huang and Haokun Chen and Guancheng Wan and Mang Ye and Xun Xiao and Hinrich Schuetze and Volker Tresp and Yunpu Ma},
      year={2025},
      eprint={2505.13408},
      archivePrefix={arXiv},
      primaryClass={cs.AI},
      url={https://arxiv.org/abs/2505.13408}, 
}

@inproceedings{dunefsky-etal-2024-transcoders,
author = {Dunefsky, Jacob and Chlenski, Philippe and Nanda, Neel},
title = {Transcoders find interpretable LLM feature circuits},
year = {2025},
isbn = {9798331314385},
publisher = {Curran Associates Inc.},
address = {Red Hook, NY, USA},
booktitle = {Proceedings of the 38th International Conference on Neural Information Processing Systems},
articleno = {768},
numpages = {36},
location = {Vancouver, BC, Canada},
series = {NIPS '24}
}

@inproceedings{vanswani-etal-2017-attention,
author = {Vaswani, Ashish and Shazeer, Noam and Parmar, Niki and Uszkoreit, Jakob and Jones, Llion and Gomez, Aidan N. and Kaiser, \L{}ukasz and Polosukhin, Illia},
title = {Attention is all you need},
year = {2017},
isbn = {9781510860964},
publisher = {Curran Associates Inc.},
address = {Red Hook, NY, USA},
booktitle = {Proceedings of the 31st International Conference on Neural Information Processing Systems},
pages = {6000–6010},
numpages = {11},
location = {Long Beach, California, USA},
series = {NIPS'17}
}

@inproceedings{
lightman2024lets,
title={Let's Verify Step by Step},
author={Hunter Lightman and Vineet Kosaraju and Yuri Burda and Harrison Edwards and Bowen Baker and Teddy Lee and Jan Leike and John Schulman and Ilya Sutskever and Karl Cobbe},
booktitle={The Twelfth International Conference on Learning Representations},
year={2024},
url={https://openreview.net/forum?id=v8L0pN6EOi}
}

@article{elhage2021mathematical,
   title={A Mathematical Framework for Transformer Circuits},
   author={Elhage, Nelson and Nanda, Neel and Olsson, Catherine and Henighan, Tom and Joseph, Nicholas and others},
   year={2021},
   journal={Transformer Circuits Thread},
   url={https://transformer-circuits.pub/2021/framework/index.html}
}

@misc{snell2024scaling,
      title={Scaling LLM Test-Time Compute Optimally can be More Effective than Scaling Model Parameters}, 
      author={Charlie Snell and Jaehoon Lee and Kelvin Xu and Aviral Kumar},
      year={2024},
      eprint={2408.03314},
      archivePrefix={arXiv},
      primaryClass={cs.LG},
      url={https://arxiv.org/abs/2408.03314}, 
}

@misc{tutek2025measuring,
      title={Measuring Chain of Thought Faithfulness by Unlearning Reasoning Steps}, 
      author={Martin Tutek and Fateme Hashemi Chaleshtori and Ana Marasović and Yonatan Belinkov},
      year={2025},
      eprint={2502.14829},
      archivePrefix={arXiv},
      primaryClass={cs.CL},
      url={https://arxiv.org/abs/2502.14829}, 
}

@inproceedings{lyu2023faithful,
  title={Faithful Chain-of-Thought Reasoning},
  author={Lyu, Qing and Havaldar, Shreya and Stein, Adam and Zhang, Li and Rao, Delip and Wong, Eric and Apidianaki, Marianna and Callison-Burch, Chris},
  booktitle={Proceedings of the 13th International Joint Conference on Natural Language Processing and the 3rd Conference of the Asia-Pacific Chapter of the Association for Computational Linguistics (Volume 1: Long Papers)},
  pages={305--329},
  year={2023}
}

@inproceedings{tyen2024llms,
  title={LLMs cannot find reasoning errors, but can correct them given the error location},
  author={Tyen, Gladys and Mansoor, Hassan and C{\u{a}}rbune, Victor and Chen, Yuanzhu Peter and Mak, Tony},
  booktitle={Findings of the Association for Computational Linguistics ACL 2024},
  pages={13894--13908},
  year={2024}
}

@misc{ji2025calibrating,
      title={Calibrating Verbal Uncertainty as a Linear Feature to Reduce Hallucinations}, 
      author={Ziwei Ji and Lei Yu and Yeskendir Koishekenov and Yejin Bang and Anthony Hartshorn and Alan Schelten and Cheng Zhang and Pascale Fung and Nicola Cancedda},
      year={2025},
      eprint={2503.14477},
      archivePrefix={arXiv},
      primaryClass={cs.CL},
      url={https://arxiv.org/abs/2503.14477}, 
}

@misc{lanham2023measuring,
      title={Measuring Faithfulness in Chain-of-Thought Reasoning}, 
      author={Tamera Lanham and Anna Chen and Ansh Radhakrishnan and others},
      year={2023},
      eprint={2307.13702},
      archivePrefix={arXiv},
      primaryClass={cs.AI},
      url={https://arxiv.org/abs/2307.13702}, 
}

@article{
bentham2024chain,
title={Chain-of-Thought Unfaithfulness as Disguised Accuracy},
author={Oliver Bentham and Nathan Stringham and Ana Marasovic},
journal={Transactions on Machine Learning Research},
issn={2835-8856},
year={2024},
url={https://openreview.net/forum?id=ydcrP55u2e},
note={Reproducibility Certification}
}

@inproceedings{xie2024calibrating,
author = {Xie, Zhihui and Guo, Jizhou and Yu, Tong and Li, Shuai},
title = {Calibrating reasoning in language models with internal consistency},
year = {2025},
isbn = {9798331314385},
publisher = {Curran Associates Inc.},
address = {Red Hook, NY, USA},
booktitle = {Proceedings of the 38th International Conference on Neural Information Processing Systems},
articleno = {3647},
numpages = {30},
location = {Vancouver, BC, Canada},
series = {NIPS '24}
}

@misc{bogdan2025thought,
      title={Thought Anchors: Which LLM Reasoning Steps Matter?}, 
      author={Paul C. Bogdan and Uzay Macar and Neel Nanda and Arthur Conmy},
      year={2025},
      eprint={2506.19143},
      archivePrefix={arXiv},
      primaryClass={cs.LG},
      url={https://arxiv.org/abs/2506.19143}, 
}

@misc{korbak2025chain,
      title={Chain of Thought Monitorability: A New and Fragile Opportunity for AI Safety}, 
      author={Tomek Korbak and Mikita Balesni and Elizabeth Barnes and Yoshua Bengio and Joe Benton and Joseph Bloom and Mark Chen and Alan Cooney and Allan Dafoe and Anca Dragan and others},
      year={2025},
      eprint={2507.11473},
      archivePrefix={arXiv},
      primaryClass={cs.AI},
      url={https://arxiv.org/abs/2507.11473}, 
}

@misc{ward2025reasoning,
      title={Reasoning-Finetuning Repurposes Latent Representations in Base Models}, 
      author={Jake Ward and Chuqiao Lin and Constantin Venhoff and Neel Nanda},
      year={2025},
      eprint={2507.12638},
      archivePrefix={arXiv},
      primaryClass={cs.LG},
      url={https://arxiv.org/abs/2507.12638}, 
}

@inproceedings{
guan2025rstarmath,
title={rStar-Math: Small {LLM}s Can Master Math Reasoning with Self-Evolved Deep Thinking},
author={Xinyu Guan and Li Lyna Zhang and Yifei Liu and Ning Shang and Youran Sun and Yi Zhu and Fan Yang and Mao Yang},
booktitle={Forty-second International Conference on Machine Learning},
year={2025},
url={https://openreview.net/forum?id=5zwF1GizFa}
}

@inproceedings{
gao2024scaling,
title={Scaling and evaluating sparse autoencoders},
author={Leo Gao and Tom Dupre la Tour and Henk Tillman and Gabriel Goh and Rajan Troll and Alec Radford and Ilya Sutskever and Jan Leike and Jeffrey Wu},
booktitle={The Thirteenth International Conference on Learning Representations},
year={2025},
url={https://openreview.net/forum?id=tcsZt9ZNKD}
}

@inproceedings{zhao2024steering,
    title = "Steering Knowledge Selection Behaviours in {LLM}s via {SAE}-Based Representation Engineering",
    author = "Zhao, Yu  and
      Devoto, Alessio  and
      Hong, Giwon  and
      Du, Xiaotang  and
      Gema, Aryo Pradipta  and
      Wang, Hongru  and
      He, Xuanli  and
      Wong, Kam-Fai  and
      Minervini, Pasquale",
    editor = "Chiruzzo, Luis  and
      Ritter, Alan  and
      Wang, Lu",
    booktitle = "Proceedings of the 2025 Conference of the Nations of the Americas Chapter of the Association for Computational Linguistics: Human Language Technologies (Volume 1: Long Papers)",
    month = apr,
    year = "2025",
    address = "Albuquerque, New Mexico",
    publisher = "Association for Computational Linguistics",
    url = "https://aclanthology.org/2025.naacl-long.264/",
    doi = "10.18653/v1/2025.naacl-long.264",
    pages = "5117--5136",
    ISBN = "979-8-89176-189-6"
}

@inproceedings{lieberum-etal-2024-gemma,
    title = "Gemma Scope: Open Sparse Autoencoders Everywhere All At Once on Gemma 2",
    author = "Lieberum, Tom  and
      Rajamanoharan, Senthooran  and
      Conmy, Arthur  and
      Smith, Lewis  and
      Sonnerat, Nicolas  and
      Varma, Vikrant  and
      Kramar, Janos  and
      Dragan, Anca  and
      Shah, Rohin  and
      Nanda, Neel",
    editor = "Belinkov, Yonatan  and
      Kim, Najoung  and
      Jumelet, Jaap  and
      Mohebbi, Hosein  and
      Mueller, Aaron  and
      Chen, Hanjie",
    booktitle = "Proceedings of the 7th BlackboxNLP Workshop: Analyzing and Interpreting Neural Networks for NLP",
    month = nov,
    year = "2024",
    address = "Miami, Florida, US",
    publisher = "Association for Computational Linguistics",
    url = "https://aclanthology.org/2024.blackboxnlp-1.19/",
    doi = "10.18653/v1/2024.blackboxnlp-1.19",
    pages = "278--300"
}

@misc{shih2023longhorizontemperaturescaling,
      title={Long Horizon Temperature Scaling}, 
      author={Andy Shih and Dorsa Sadigh and Stefano Ermon},
      year={2023},
      eprint={2302.03686},
      archivePrefix={arXiv},
      primaryClass={cs.LG},
      url={https://arxiv.org/abs/2302.03686}, 
}

@inproceedings{liu-etal-2020-energy,
author = {Liu, Weitang and Wang, Xiaoyun and Owens, John D. and Li, Yixuan},
title = {Energy-based out-of-distribution detection},
year = {2020},
isbn = {9781713829546},
publisher = {Curran Associates Inc.},
address = {Red Hook, NY, USA},
booktitle = {Proceedings of the 34th International Conference on Neural Information Processing Systems},
articleno = {1802},
numpages = {12},
location = {Vancouver, BC, Canada},
series = {NIPS '20}
}

@article{llama3modelcard,
title={Llama 3 Model Card},
author={AI@Meta},
year={2024},
url = {https://github.com/meta-llama/llama3/blob/main/MODEL_CARD.md}
}

@inproceedings{
weber2024redpajama,
title={RedPajama: an Open Dataset for Training Large Language Models},
author={Maurice Weber and Daniel Y Fu and Quentin Gregory Anthony and Yonatan Oren and Shane Adams and Anton Alexandrov and Xiaozhong Lyu and Huu Nguyen and Xiaozhe Yao and Virginia Adams and Ben Athiwaratkun and Rahul Chalamala and Kezhen Chen and Max Ryabinin and Tri Dao and Percy Liang and Christopher Re and Irina Rish and Ce Zhang},
booktitle={The Thirty-eight Conference on Neural Information Processing Systems Datasets and Benchmarks Track},
year={2024},
url={https://openreview.net/forum?id=lnuXaRpwvw}
}

@misc{yang2025diversitydrivendataselectionlanguage,
      title={Diversity-driven Data Selection for Language Model Tuning through Sparse Autoencoder}, 
      author={Xianjun Yang and Shaoliang Nie and Lijuan Liu and Suchin Gururangan and Ujjwal Karn and Rui Hou and Madian Khabsa and Yuning Mao},
      year={2025},
      eprint={2502.14050},
      archivePrefix={arXiv},
      primaryClass={cs.CL},
      url={https://arxiv.org/abs/2502.14050}, 
}

@misc{circuit-tracer,
  author = {Hanna, Michael and Piotrowski, Mateusz and Lindsey, Jack and Ameisen, Emmanuel},
  title = {circuit-tracer},
  howpublished = {\url{https://github.com/safety-research/circuit-tracer}},
  note = {The first two authors contributed equally and are listed alphabetically.},
  year = {2025}
}

@inproceedings{
arcuschin2025chainofthought,
title={Chain-of-Thought Reasoning in the Wild is not Always Faithful},
author={Iv{\'a}n Arcuschin and Jett Janiak and Robert Krzyzanowski and Senthooran Rajamanoharan and Neel Nanda and Arthur Conmy},
booktitle={Workshop on Reasoning and Planning for Large Language Models},
year={2025},
url={https://openreview.net/forum?id=L8094Whth0}
}

@inproceedings{turpin-etal-2023-language,
author = {Turpin, Miles and Michael, Julian and Perez, Ethan and Bowman, Samuel R.},
title = {Language models don't always say what they think: unfaithful explanations in chain-of-thought prompting},
year = {2023},
publisher = {Curran Associates Inc.},
address = {Red Hook, NY, USA},
booktitle = {Proceedings of the 37th International Conference on Neural Information Processing Systems},
articleno = {3275},
numpages = {14},
location = {New Orleans, LA, USA},
series = {NIPS '23}
}

@article{lindsey2025biology,
  author={Lindsey, Jack and Gurnee, Wes and Ameisen, Emmanuel and Chen, Brian and Pearce, Adam and Turner, Nicholas L. and Citro, Craig and Abrahams, David and Carter, Shan and Hosmer, Basil and Marcus, Jonathan and Sklar, Michael and Templeton, Adly and Bricken, Trenton and McDougall, Callum and Cunningham, Hoagy and Henighan, Thomas and Jermyn, Adam and Jones, Andy and Persic, Andrew and Qi, Zhenyi and Thompson, T. Ben and Zimmerman, Sam and Rivoire, Kelley and Conerly, Thomas and Olah, Chris and Batson, Joshua},
  title={On the Biology of a Large Language Model},
  journal={Transformer Circuits Thread},
  year={2025},
  url={https://transformer-circuits.pub/2025/attribution-graphs/biology.html}
}

@misc{chen2025reasoningmodelsdontsay,
      title={Reasoning Models Don't Always Say What They Think}, 
      author={Yanda Chen and Joe Benton and Ansh Radhakrishnan and Jonathan Uesato and Carson Denison and John Schulman and Arushi Somani and Peter Hase and Misha Wagner and Fabien Roger and Vlad Mikulik and Samuel R. Bowman and Jan Leike and Jared Kaplan and Ethan Perez},
      year={2025},
      eprint={2505.05410},
      archivePrefix={arXiv},
      primaryClass={cs.CL},
      url={https://arxiv.org/abs/2505.05410}, 
}

@inproceedings{
zhang2025reasoning,
title={Reasoning Models Know When They{\textquoteright}re Right: Probing Hidden States for Self-Verification},
author={Anqi Zhang and Yulin Chen and Jane Pan and Chen Zhao and Aurojit Panda and Jinyang Li and He He},
booktitle={Second Conference on Language Modeling},
year={2025},
url={https://openreview.net/forum?id=O6I0Av7683}
}

@inproceedings{wang-etal-2025-chain,
    title = "Chain-of-Probe: Examining the Necessity and Accuracy of {C}o{T} Step-by-Step",
    author = "Wang, Zezhong  and
      Zeng, Xingshan  and
      Liu, Weiwen  and
      Wang, Yufei  and
      Li, Liangyou  and
      Wang, Yasheng  and
      Shang, Lifeng  and
      Jiang, Xin  and
      Liu, Qun  and
      Wong, Kam-Fai",
    editor = "Chiruzzo, Luis  and
      Ritter, Alan  and
      Wang, Lu",
    booktitle = "Findings of the Association for Computational Linguistics: NAACL 2025",
    month = apr,
    year = "2025",
    address = "Albuquerque, New Mexico",
    publisher = "Association for Computational Linguistics",
    url = "https://aclanthology.org/2025.findings-naacl.140/",
    doi = "10.18653/v1/2025.findings-naacl.140",
    pages = "2586--2606",
    ISBN = "979-8-89176-195-7"
}

@inproceedings{afzal-etal-2025-knowing,
    title = "Knowing Before Saying: {LLM} Representations Encode Information About Chain-of-Thought Success Before Completion",
    author = "Afzal, Anum  and
      Matthes, Florian  and
      Chechik, Gal  and
      Ziser, Yftah",
    editor = "Che, Wanxiang  and
      Nabende, Joyce  and
      Shutova, Ekaterina  and
      Pilehvar, Mohammad Taher",
    booktitle = "Findings of the Association for Computational Linguistics: ACL 2025",
    month = jul,
    year = "2025",
    address = "Vienna, Austria",
    publisher = "Association for Computational Linguistics",
    url = "https://aclanthology.org/2025.findings-acl.662/",
    doi = "10.18653/v1/2025.findings-acl.662",
    pages = "12791--12806",
    ISBN = "979-8-89176-256-5"
}

@misc{baker2025monitoringreasoningmodelsmisbehavior,
      title={Monitoring Reasoning Models for Misbehavior and the Risks of Promoting Obfuscation}, 
      author={Bowen Baker and Joost Huizinga and Leo Gao and Zehao Dou and Melody Y. Guan and Aleksander Madry and Wojciech Zaremba and Jakub Pachocki and David Farhi},
      year={2025},
      eprint={2503.11926},
      archivePrefix={arXiv},
      primaryClass={cs.AI},
      url={https://arxiv.org/abs/2503.11926}, 
}

@misc{deepseekai2025deepseekr1,
      title={DeepSeek-R1: Incentivizing Reasoning Capability in LLMs via Reinforcement Learning}, 
      author={DeepSeek-AI and others},
      year={2025},
      eprint={2501.12948},
      archivePrefix={arXiv},
      primaryClass={cs.CL},
      url={https://arxiv.org/abs/2501.12948}, 
}

@misc{openai2024openaio1card,
      title={OpenAI o1 System Card}, 
      author={OpenAI and others},
      year={2024},
      eprint={2412.16720},
      archivePrefix={arXiv},
      primaryClass={cs.AI},
      url={https://arxiv.org/abs/2412.16720}, 
}

@misc{li2025thinkingfail,
      title={When Thinking Fails: The Pitfalls of Reasoning for Instruction-Following in LLMs}, 
      author={Xiaomin Li and Zhou Yu and Zhiwei Zhang and Xupeng Chen and Ziji Zhang and Yingying Zhuang and Narayanan Sadagopan and Anurag Beniwal},
      year={2025},
      eprint={2505.11423},
      archivePrefix={arXiv},
      primaryClass={cs.CL},
      url={https://arxiv.org/abs/2505.11423}, 
}

@inproceedings{jacovi-etal-2024-chain,
    title = "A Chain-of-Thought Is as Strong as Its Weakest Link: A Benchmark for Verifiers of Reasoning Chains",
    author = "Jacovi, Alon  and
      Bitton, Yonatan  and
      Bohnet, Bernd  and
      Herzig, Jonathan  and
      Honovich, Or  and
      Tseng, Michael  and
      Collins, Michael  and
      Aharoni, Roee  and
      Geva, Mor",
    editor = "Ku, Lun-Wei  and
      Martins, Andre  and
      Srikumar, Vivek",
    booktitle = "Proceedings of the 62nd Annual Meeting of the Association for Computational Linguistics (Volume 1: Long Papers)",
    month = aug,
    year = "2024",
    address = "Bangkok, Thailand",
    publisher = "Association for Computational Linguistics",
    url = "https://aclanthology.org/2024.acl-long.254/",
    doi = "10.18653/v1/2024.acl-long.254",
    pages = "4615--4634"
}

@inproceedings{wang-etal-2024-math,
    title = "Math-Shepherd: Verify and Reinforce {LLM}s Step-by-step without Human Annotations",
    author = "Wang, Peiyi  and
      Li, Lei  and
      Shao, Zhihong  and
      Xu, Runxin  and
      Dai, Damai  and
      Li, Yifei  and
      Chen, Deli  and
      Wu, Yu  and
      Sui, Zhifang",
    editor = "Ku, Lun-Wei  and
      Martins, Andre  and
      Srikumar, Vivek",
    booktitle = "Proceedings of the 62nd Annual Meeting of the Association for Computational Linguistics (Volume 1: Long Papers)",
    month = aug,
    year = "2024",
    address = "Bangkok, Thailand",
    publisher = "Association for Computational Linguistics",
    url = "https://aclanthology.org/2024.acl-long.510/",
    doi = "10.18653/v1/2024.acl-long.510",
    pages = "9426--9439"
}

@misc{kissane2024saes,
  author= {Connor Kissane and Robert Krzyzanowski and Arthur Conmy and Neel Nanda},
  url = {https://www.alignmentforum.org/posts/fmwk6qxrpW8d4jvbd/saes-usually-transfer-between-base-and-chat-models},
  year = {2024},
  howpublished = {Alignment Forum},
  title = {SAEs (usually) Transfer Between Base and Chat Models},
}

@inproceedings{
zheng2024lmsyschatm,
title={{LMSYS}-Chat-1M: A Large-Scale Real-World {LLM} Conversation Dataset},
author={Lianmin Zheng and Wei-Lin Chiang and Ying Sheng and Tianle Li and Siyuan Zhuang and Zhanghao Wu and Yonghao Zhuang and Zhuohan Li and Zi Lin and Eric Xing and Joseph E. Gonzalez and Ion Stoica and Hao Zhang},
booktitle={The Twelfth International Conference on Learning Representations},
year={2024},
url={https://openreview.net/forum?id=BOfDKxfwt0}
}

@article{scikit-learn,
  title={Scikit-learn: Machine Learning in {P}ython},
  author={Pedregosa, F. and Varoquaux, G. and Gramfort, A. and Michel, V.
          and Thirion, B. and Grisel, O. and Blondel, M. and Prettenhofer, P.
          and Weiss, R. and Dubourg, V. and Vanderplas, J. and Passos, A. and
          Cournapeau, D. and Brucher, M. and Perrot, M. and Duchesnay, E.},
  journal={Journal of Machine Learning Research},
  volume={12},
  pages={2825--2830},
  year={2011}
}

@misc{lee2025geometry,
      title={The Geometry of Self-Verification in a Task-Specific Reasoning Model}, 
      author={Andrew Lee and Lihao Sun and Chris Wendler and Fernanda Viégas and Martin Wattenberg},
      year={2025},
      eprint={2504.14379},
      archivePrefix={arXiv},
      primaryClass={cs.AI},
      url={https://arxiv.org/abs/2504.14379}, 
}

@misc{venhoff2025basemodelsknowreason,
      title={Base Models Know How to Reason, Thinking Models Learn When}, 
      author={Constantin Venhoff and Iván Arcuschin and Philip Torr and Arthur Conmy and Neel Nanda},
      year={2025},
      eprint={2510.07364},
      archivePrefix={arXiv},
      primaryClass={cs.AI},
      url={https://arxiv.org/abs/2510.07364}, 
}

\clearpage
\newpage
\beginappendix

\section{Additional Details on Datasets}
\label{sec:appendix_dataset}
Here we provide a detailed description of our dataset construction, our labeling protocol, and the final dataset statistics.

\subsection{Synthetic Dataset Construction}
\label{sec:dataset_construction_appendix}
To create a controlled environment for studying reasoning failures, we procedurally generated two synthetic datasets: {\bf Boolean} and {\bf Arithmetic}. For each, we first generated a ground-truth expression, then prompted our base model (Llama 3.1 8B Instruct) to produce a Chain-of-Thought solution towards solving the expression. We provide the prompt template used to generate CoT in Table~\ref{tab:prompts_cot_generation}. Once the CoT is generated, we split them into steps using regular expression.

\begin{table}[h!]
\footnotesize
\centering

\begin{tabular}{@{}ll@{}}
\toprule
{\bf Dataset} & {\bf Llama 3.1 8B Instruct Prompt Template} \\
\midrule

{\bf Boolean} &
\parbox[t]{0.8\textwidth}{
\texttt{<|begin\_of\_text|><|start\_header\_id|>system<|end\_header\_id|>}\\[0.5em]
Evaluate the boolean expression below.\\[0.5em]
\texttt{<|eot\_id|><|start\_header\_id|>user<|end\_header\_id|>}\\[0.5em]
\textit{\{boolean\_expression\}}\\[0.5em]
\texttt{<|eot\_id|><|start\_header\_id|>assistant<|end\_header\_id|>}
} \\
\midrule

{\bf Arithmetic} &
\parbox[t]{0.8\textwidth}{
\texttt{<|begin\_of\_text|><|start\_header\_id|>system<|end\_header\_id|>}\\[0.5em]
Evaluate the arithmetic expression below.\\[0.5em]
\texttt{<|eot\_id|><|start\_header\_id|>user<|end\_header\_id|>}\\[0.5em]
\textit{\{arithmetic\_expression\}}\\[0.5em]
\texttt{<|eot\_id|><|start\_header\_id|>assistant<|end\_header\_id|>}
} \\
\midrule

{\bf GSM8K} &
\parbox[t]{0.8\textwidth}{
\texttt{<|begin\_of\_text|><|start\_header\_id|>system<|end\_header\_id|>}\\[0.5em]
Solve the following math problem step by step.\\[0.5em]
\texttt{<|eot\_id|><|start\_header\_id|>user<|end\_header\_id|>}\\[0.5em]
\textit{\{gsm8k\_question\}}\\[0.5em]
\texttt{<|eot\_id|><|start\_header\_id|>assistant<|end\_header\_id|>}
} \\
\bottomrule
\end{tabular}
\caption{Prompts used for CoT generation across the three datasets. Placeholders for dynamic content are shown in \textit{italics}.}
\label{tab:prompts_cot_generation}
\end{table}

\paragraph{\bf Expression Generation.}
The complexity of the expressions was controlled by the number of operators, $n \in \{3, 5, 7, 10\}$. For the Arithmetic task, we generated expressions with single-digit integers and the operators $\{+, -, \times\}$. For the Boolean task, operands were simply $\{\texttt{True}, \texttt{False}\}$, and possible operators are $\{\texttt{and}, \texttt{or}, \texttt{not}\}$. We generated 10,000 unique expressions for each difficulty level, with the exception of the Boolean task at $n=3$, where the number of unique expressions is inherently limited. Table~\ref{tab:synthetic_dataset_examples} shows examples from these datasets with their associated CoTs generated by Llama 3.1 8B Instruct model.

\begin{table}[h!]
\centering
\footnotesize
\begin{tabular}{p{0.15\textwidth} | p{0.75\textwidth}}
\toprule
{\bf Task} & {\bf Example} \\
\midrule
{\bf Boolean} ($n=5$) & 
    {\bf Expression:} 
    {\ttfamily (((True or True) and (True and True)) or (True and False))}
    \vspace{1em}\par 
    {\bf Generated CoT:} To evaluate the boolean expression, we need to follow the order of operations (PEMDAS): \par
    1. Evaluate the expressions inside the innermost parentheses: \par
    \quad * {\ttfamily (True or True) = True} \par
    \quad * {\ttfamily (True and True) = True} \par
    \vspace{0.5em}
    2. Now the expression becomes: \par
    \quad * {\ttfamily (True and True) or (True and False)} \par
    \vspace{0.5em}
    3. Evaluate the expressions inside the parentheses: \par
    \quad * {\ttfamily (True and True) = True} \par
    \quad * {\ttfamily (True and False) = False} \par
    \vspace{0.5em}
    4. Now the expression becomes: \par
    \quad * {\ttfamily True or False} \par
    \vspace{0.5em}
    5. Evaluate the final expression: \par
    \quad * {\ttfamily True or False = True} \par
    \vspace{1em}
    The final answer is {\ttfamily True}.
    \\ 
\midrule
{\bf Arithmetic} ($n=7$) & 
    {\bf Expression:} 
    {\ttfamily ((((-3)+(-6))*(9*6))+(-4))}
    \vspace{1em}\par
    {\bf Generated CoT:} To evaluate this expression, we need to follow the order of operations (PEMDAS): \par
    \vspace{0.5em}
    1. Evaluate the expressions inside the innermost parentheses: \par
    \quad * {\ttfamily (-3) + (-6) = -9} \par
    \vspace{0.5em}
    2. Multiply 9 and 6: \par
    \quad * {\ttfamily 9 * 6 = 54} \par
    \vspace{0.5em}
    3. Multiply -9 and 54: \par
    \quad * {\ttfamily -9 * 54 = -486} \par
    \vspace{0.5em}
    4. Add -4 to -486: \par
    \quad * {\ttfamily -486 + (-4) = -490} \par
    \vspace{1em}
    The final answer is -490.
    \\
\bottomrule
\end{tabular}
\caption{Examples of generated expressions and the corresponding CoT generated by our base model for the synthetic datasets.}
\label{tab:synthetic_dataset_examples}
\end{table}

\subsection{Labeling Protocol}
\label{sec:labeling_appendix}
Our primary goal was to create labels of the highest possible quality. For the synthetic datasets, we developed a rigorous two-pronged strategy that leverages both a powerful LLM judge and direct programmatic verification, keeping only the annotations where both methods agree.

\paragraph{\bf Method 1: LLM-as-a-Judge.}
Our first annotation method uses a powerful, external LLM---Llama 3.3 70B Instruct---as an expert judge. For each reasoning step, the judge model was provided with the full problem context and the generated step, and was prompted to output a binary correctness label and a justification. This method was used for all three datasets including the GSM8K dataset. We provide the prompts used for the judge in Table~\ref{tab:judge_prompt_boolean} (Boolean), Table~\ref{tab:judge_prompt_arithmetic} (Arithmetic), and Table~\ref{tab:judge_prompt_gsm8k} (GSM8K). 

\paragraph{\bf Method 2: Programmatic State Verification (Synthetic Datasets only).}
For the synthetic tasks, we developed a programmatic method to verify the logical integrity of the entire reasoning trace. This goes beyond simply checking the correctness of a single calculation. After each reasoning step $i$ generated by the model, we use a specially crafted prompt to ask the model to output the \textit{current, simplified state} of the original expression.

For example, if the original expression is \texttt{(3+5)*2} and the model's first step is \texttt{1. 3+5=8}, we then prompt it: ``Now the original expression becomes: ''. The model is expected to return the reduced expression \texttt{8*2}. We then programmatically evaluate the ground-truth value of both the original expression and this new, simplified expression returned by the model. A step is labeled `correct' only if the two values are identical. If at any point the value of the simplified expression diverges from the ground-truth value of the original, that step is labeled `incorrect'. While occasionally the model outputs a reduced expression which evaluates to the same value despite being incorrect, this method filters a significant amount of errors.

\paragraph{\bf Final Label Agreement.}
To create our final, high-fidelity label set for the synthetic tasks, we took the intersection of the labels from both methods. That is, a reasoning step was only included in our final dataset if \textit{both} the LLM-as-a-Judge and the programmatic verifier agreed on its label. This strict agreement protocol ensures an exceptionally clean dataset by filtering out ambiguous cases or potential errors from either annotation method.\footnote{While this significantly increases our confidence in the label quality, it also has the effect of making the class distribution more imbalanced, as ambiguous incorrect cases are more likely to be filtered out.}

\begin{table}[t]
\footnotesize
\centering
\begin{tabular}{@{}p{0.95\textwidth}@{}}
\toprule
{\bf Llama 3.3 70B Instruct Prompt Template} \\
\midrule
\texttt{<|begin\_of\_text|><|start\_header\_id|>system<|end\_header\_id|>}\\[0.5em]
You are an expert in logical reasoning and boolean algebra. You evaluate the correctness of reasoning steps in boolean expression evaluation with high precision.\\[0.5em]
\texttt{<|eot\_id|><|start\_header\_id|>user<|end\_header\_id|>}\\[0.5em]
Evaluate this reasoning step for logical correctness:\\[1em]
Original Boolean Expression: \textit{\{original\_expression\}} \\
Correct Truth Value: \textit{\{correct\_value\}}\\[1em]
Context (previous steps):\\
\textit{\{context\}}\\[1em]
Step to evaluate: \textit{\{step\}}\\[1em]
Evaluation criteria:\\
- Is the boolean operation applied correctly?\\
- Does the step follow proper order of operations?\\
- Are the truth values computed accurately?\\
- Is the reasoning logically sound?\\[1em]
Respond with exactly one of the following:\\
- CORRECT: if the step is logically sound and mathematically accurate\\
- INCORRECT: if the step contains logical errors, mathematical mistakes, or invalid reasoning\\[1em]
Your response should start with either ``CORRECT'' or ``INCORRECT'' followed by a brief explanation.\\[0.5em]
\texttt{<|eot\_id|><|start\_header\_id|>assistant<|end\_header\_id|>}
\\ \bottomrule
\end{tabular}
\caption{Prompt used for step-level annotation by the Llama 3.3 70B Instruct judge model on the Synthetic Boolean dataset. Placeholders for dynamic content are shown in \textit{italics}.}
\label{tab:judge_prompt_boolean}
\end{table}

\begin{table}[t]
\footnotesize
\centering 
\begin{tabular}{@{}p{0.95\textwidth}@{}}
\toprule
{\bf Llama 3.3 70B Instruct Prompt Template} \\
\midrule
\texttt{<|begin\_of\_text|><|start\_header\_id|>system<|end\_header\_id|>}\\[0.5em]
You are an expert in mathematical reasoning and arithmetic operations. You evaluate the correctness of reasoning steps in arithmetic expression evaluation with high precision.\\[0.5em]
\texttt{<|eot\_id|><|start\_header\_id|>user<|end\_header\_id|>}\\[0.5em]
Evaluate this reasoning step for mathematical correctness:\\[1em]
Original Arithmetic Expression: \textit{\{original\_expression\}} \\
Correct Value: \textit{\{correct\_value\}}\\[1em]
Context (previous steps):\\
\textit{\{context\}}\\[1em]
Step to evaluate: \textit{\{step\}}\\[1em]
Evaluation criteria:\\
- Are the arithmetic operations applied correctly?\\
- Does the step follow proper order of operations (PEMDAS/BODMAS)?\\
- Are the numerical computations accurate?\\
- Is the mathematical reasoning sound?\\[1em]
Respond with exactly one of the following:\\
- CORRECT: if the step is mathematically sound and computationally accurate\\
- INCORRECT: if the step contains mathematical errors, computational mistakes, or invalid reasoning\\[1em]
Your response should start with either ``CORRECT'' or ``INCORRECT'' followed by a brief explanation.\\[0.5em]
\texttt{<|eot\_id|><|start\_header\_id|>assistant<|end\_header\_id|>}
\\ \bottomrule
\end{tabular}
\caption{Prompt used for step-level annotation by the Llama 3.3 70B Instruct judge model on the Synthetic Arithmetic dataset. Placeholders for dynamic content are shown in \textit{italics}.}
\label{tab:judge_prompt_arithmetic}
\end{table}

\begin{table}[t]
\footnotesize
\centering
\begin{tabular}{@{}p{0.95\textwidth}@{}}
\toprule
{\bf Llama 3.3 70B Instruct Prompt Template} \\
\midrule
\texttt{<|begin\_of\_text|><|start\_header\_id|>system<|end\_header\_id|>}\\[0.5em]
You are an expert in mathematical word problems and quantitative reasoning. Your purpose is to evaluate a single reasoning step taken to solve a multi-step word problem. You must be precise, focusing only on the provided step and its relationship to the problem and previously established facts.\\[0.5em]
\texttt{<|eot\_id|><|start\_header\_id|>user<|end\_header\_id|>}\\[0.5em]
Your task is to evaluate the provided reasoning step for logical and mathematical correctness.\\[1em]
Original Math Problem: \textit{\{original\_question\}}\\
Correct Final Answer: \textit{\{correct\_value\}}\\[1em]
Context (previous steps):\\
\textit{\{context\}}\\[1em]
Step to evaluate: \textit{\{step\}}\\[1em]
Evaluation criteria:\\
- Does the step correctly extract and interpret information from the `Original Problem' or the `Context'?\\
- Is it using the right numbers for the right concepts?\\
- Is the chosen mathematical operation (e.g., addition, subtraction) the correct one to achieve the step's goal, based on the narrative of the `Original Problem'?\\
- Is the arithmetic in the step performed correctly?\\
- Is the mathematical reasoning sound?\\
- Is the step logically consistent with the problem and previous steps?\\
- The following types of steps do not contain an error and must be classified as CORRECT:\\
\quad  - A simple, factually accurate restatement of information from the problem or context.\\
\quad  - A non-substantive introductory or conversational phrase (e.g., ``Let's solve this step by step'', ``First, we need to find...'').\\[2em]
Respond with exactly one of the following:\\
- CORRECT: if the step is mathematically sound and computationally accurate\\
- INCORRECT: if the step contains mathematical errors, computational mistakes, or invalid reasoning\\[1em]
Your response should start with either ``CORRECT'' or ``INCORRECT'' followed by a brief explanation.\\[0.5em]
\texttt{<|eot\_id|><|start\_header\_id|>assistant<|end\_header\_id|>}
\\ \bottomrule
\end{tabular}
\caption{Prompt used for step-level annotation by the Llama 3.3 70B Instruct judge model on the GSM8K dataset. Placeholders for dynamic content are shown in \textit{italics}.}
\label{tab:judge_prompt_gsm8k}
\end{table}

\subsection{Human Validation of LLM-as-a-Judge Labels}

To validate the quality of our LLM-as-a-Judge annotation pipeline, a subset of 100 randomly sampled Boolean and Arithmetic expressions ($\approx$700 steps) was independently annotated by four authors. Each annotator labeled half of the set, with every step covered by at least two annotators. To mitigate the rarity of incorrect steps, we upsampled the positive class. Because of the extreme class imbalance, Cohen’s Kappa ($\kappa$) can underestimate agreement, so we report both $\kappa$ and raw percentage agreement to give a fuller view of inter-annotator reliability.

The results are summarized in Table~\ref{tab:human_annotation}. The agreement among human annotators was moderate as measured by Cohen's Kappa ($\kappa = 0.42$) but high in simple agreement (87.3\%). When comparing the consensus human labels to the LLM-as-a-Judge labels, we found fair agreement by Kappa ($\kappa = 0.26$) and similarly high simple agreement (84.1\%). A qualitative review of the disagreements revealed a recurring pattern: the vast majority of discrepancies, both among humans and between humans and the LLM judge, occurred on steps that followed an initial reasoning error. This highlights the inherent ambiguity of labeling steps on a corrupted computational path and directly motivates our strict truncation policy, as detailed in the following section.

\begin{table}[h!]
\centering
\begin{tabular}{@{}lcc@{}}
\toprule
{\bf Comparison} & {\bf Cohen's Kappa ($\kappa$)} & {\bf Agreement (\%)} \\
\midrule
Human vs. Human (Inter-Annotator) & 0.42 & 87.3\% \\
Human vs. LLM-as-a-Judge & 0.26 & 84.1\% \\
\bottomrule
\end{tabular}
\caption{Inter-Annotator Agreement (IAA) statistics for the human validation study. The comparison shows moderate-to-high agreement, with lower Kappa scores reflecting the extreme class imbalance.}
\label{tab:human_annotation}
\end{table}

\subsection{Handling Error Propagation}

A key challenge in labeling CoT is handling steps that occur after the first mistake. A model can make a technically correct calculation based on a previously incorrect number or expression. This creates a fundamental ambiguity: should a step be judged on its local correctness or its global contribution to the final answer?

To illustrate this problem, consider the flawed CoT trace shown in Table~\ref{tab:error_propagation_example}. The model correctly evaluates the innermost parentheses in Step 1. However, in Step 2, it makes a critical structural error by incorrectly dropping parentheses, which alters the order of operations. The correct simplification in Step 4 should have been ``\texttt{not (False or True)}''. While subsequent steps like 5 and 7 are arithmetically valid given the new, flawed expression from Step 2, their ultimate correctness is undefined as they exist on a corrupted computational path. To avoid this ambiguity and create a clean, well-defined verification task, we adopted a strict truncation policy. For any given reasoning chain, we only retain the steps up to and including the \textit{first} incorrect step. All subsequent steps are discarded from our dataset.

\begin{table}[t]
\footnotesize
\centering
\resizebox{\textwidth}{!}{
\begin{tabular}{@{}p{0.95\textwidth}@{}}
\toprule
{\bf Original Expression:} \texttt{( not ( ( False or ( True and False ) ) or ( True or False ) ) )} \\
\midrule
{\bf Model's Generated Chain-of-Thought:} To evaluate this expression, we need to follow the order of operations (PEMDAS): \\
\quad 1. Evaluate the innermost parentheses: \texttt{(True and False) = False}. So, \texttt{(False or (True and False)) = False or False = False} \\
\quad 2. \sethlcolor{myred}\hl{Now, the expression becomes: \texttt{(not (False or (True and False)) or (True or False))}
} \\
\quad 3. Evaluate the next innermost parentheses: \texttt{(True or False) = True}. So, \texttt{(not (False or (True and False)) or (True or False)) = (not False or True) } \\
\quad 4. Now, the expression becomes: \texttt{(not False or True)} \\
\quad 5. Evaluate the NOT operator: \texttt{not False = True} \\
\quad 6. Now, the expression becomes: \texttt{True or True} \\
\quad 7. Finally, evaluate the OR operator: \texttt{True or True = True} \\
\quad Therefore, the final result of the boolean expression is: \texttt{True} \\
\bottomrule
\end{tabular}
}
\caption{An example of a flawed CoT trace demonstrating error propagation. The \sethlcolor{myred}\hl{first unrecoverable error} in Step 2 renders the correctness of all subsequent steps ambiguous.}
\label{tab:error_propagation_example}
\end{table}

\subsection{Dataset Statistics and Model Performance}
\label{app:dataset_statistics}
For GSM8K, we used the official testset split. Since its reasoning steps are expressed in natural language and are not always programmatically verifiable, we relied solely on our validated LLM-as-a-Judge pipeline for annotation. Table~\ref{tab:appendix_stats} provides the final statistics for all three datasets, including size and label distribution. Table~\ref{tab:appendix_model_perf} reports the base performance of our Llama 3.1 8B Instruct model on these tasks. For our synthetic datasets, we randomly split the data into 80\% training and 20\% testing for the subsequent classification task.

\begin{table}[h!]
\centering
\begin{tabular}{@{}l ccc@{}}
\toprule
{\bf Dataset} & {\bf Total Steps} & {\bf \% Correct} & {\bf \% Incorrect} \\
\midrule
Synthetic (Boolean) & 126,624 & 99.8\% & 0.2\% \\
Synthetic (Arithmetic) & 155,434 & 98.8\% & 1.2\% \\
GSM8K & 8,737 & 93.4\% & 6.6\% \\
\bottomrule
\end{tabular}
\caption{Final statistics of our curated datasets, showing the number of reasoning steps and the distribution of correct/incorrect labels after our full annotation and filtering process.}
\label{tab:appendix_stats}
\end{table}

\begin{table}[h!]
\centering
\begin{tabular}{@{}llc@{}}
\toprule
{\bf Dataset} & {\bf Difficulty (Operators)} & {\bf Final Answer Accuracy} \\
\midrule
\multirow{5}{*}{Synthetic (Boolean)} & n=3 & 98.4\% \\
 & n=5 & 93.27\% \\
 & n=7 & 89.4\% \\
 & n=10 & 78.43\% \\
\midrule
\multirow{5}{*}{Synthetic (Arithmetic)} & n=3 & 94.83\% \\
 & n=5 & 86.8\% \\
 & n=7 & 73.07\% \\
 & n=10 & 52.8\% \\
\midrule
GSM8K & - & 75.82\% \\
\bottomrule
\end{tabular}
\caption{End-to-end task accuracy of our base model (Llama 3.1 8B Instruct). For the synthetic datasets, we provide a fine-grained breakdown by difficulty, controlled by the number of operators ($n$).}
\label{tab:appendix_model_perf}
\end{table}

\clearpage
\section{Transcoder Training Details}
\label{sec:appendix_transcoder}

Our methodology relies on high-fidelity, sparsely activating transcoders to create an interpretable model. To this end, we trained a TopK-Transcoder for each target MLP module in the Llama 3.1 8B Instruct model. Our training protocol is designed for robustness and follows several best practices established in recent literature.

The transcoders were trained on a high-quality, 10B token subset of the RedPajama-V2 dataset~\citep{weber2024redpajama}. We pre-processed the entire training corpus by concatenating and chunking all passages into a uniform length, and we explicitly discarded all beginning-of-sequence (BOS) tokens\footnote{BOS tokens are retained when generating activations but their activations are removed afterward for training the transcoders.}, which we found to be detrimental to stable transcoder training. The transcoder architecture consists of a simple autoencoder with a single hidden layer and a ReLU activation. For each MLP layer in the base model, the transcoder is trained to take the residual stream before the MLP block as input and reconstruct the residual stream after the MLP's computation. The input dimension matches the Llama 3.1 8B's MLP hidden dimension (4096), and the latent feature dimension was set to an overcomplete basis of 131,072. We enforced sparsity structurally using a TopK mechanism, preserving only the $k=128$ largest feature activations in the forward pass.

We followed several established training techniques to improve feature quality and avoid common pitfalls \citep{gao2024scaling,yang2025diversitydrivendataselectionlanguage}. The decoder weights were normalized to have unit norm, and we did not tie the encoder and decoder weights. To prevent feature collapse, we implemented a dead neuron revival mechanism: if a feature neuron had not activated in 10 million tokens, its activation was forced with an auxiliary loss (coefficient of 1/32).

\begin{figure}[h]
    \centering
    \includegraphics[width=0.8\linewidth]{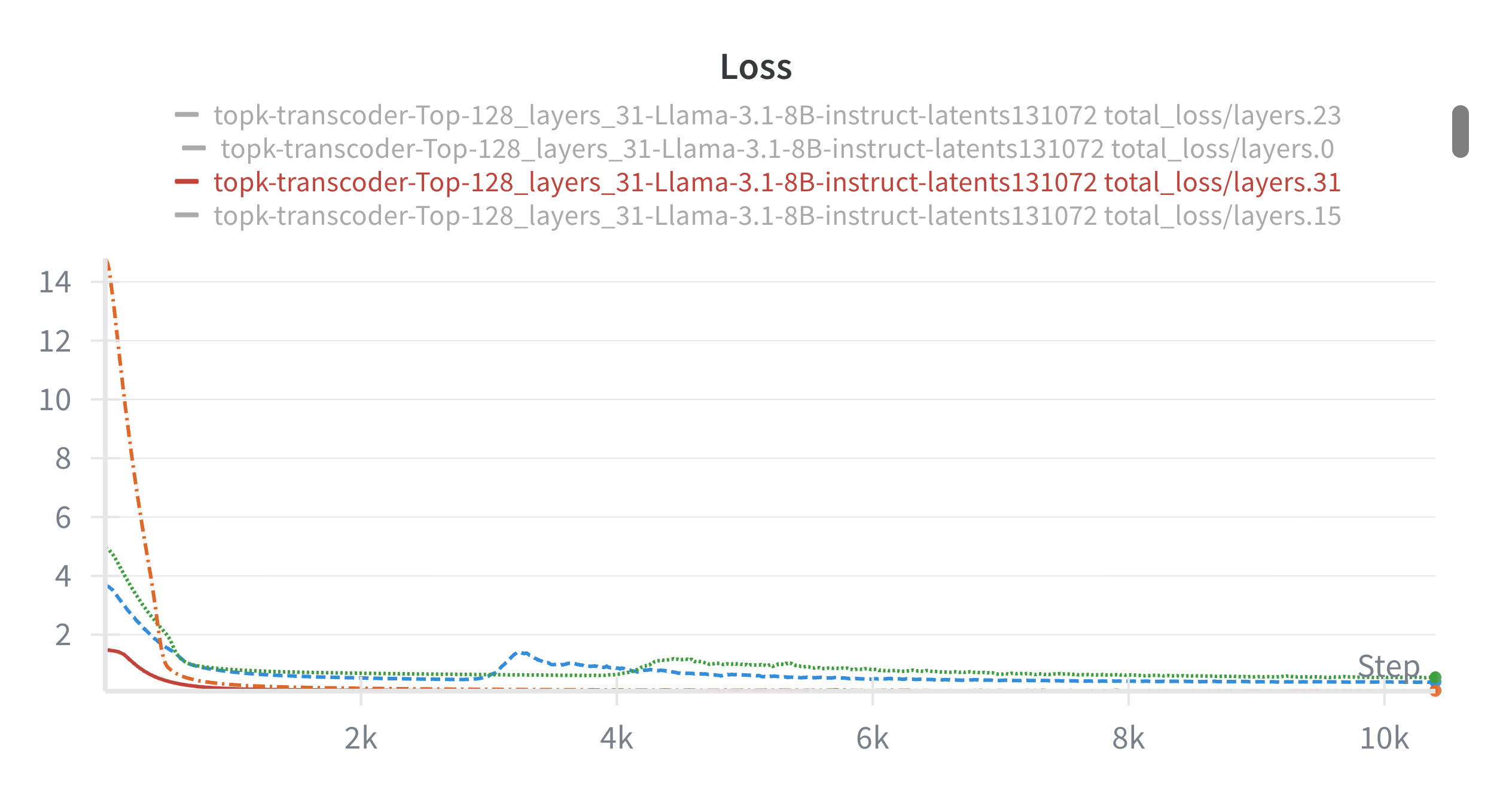} 
    \caption{
        {Transcoder Training Loss Curves.} The x-axis represents training steps. In all cases, the loss converges efficiently, generally saturating after approximately 4,000 steps.
    }
    \label{fig:transcoder_loss_curves}
\end{figure}

The transcoders were trained for 4 epochs using the AdamW optimizer. The learning rate was set to 7e-5 with a warmup ratio of 0.5. Training was conducted on 4 nodes, each with 8 Nvidia H200 GPUs, using a total batch size of 4,096. This was achieved with a per-device batch size of 32 and gradient accumulation steps. We found that the training loss generally saturated after approximately 4,000 steps, indicating efficient convergence. We show the training loss on selected layers in Figure~\ref{fig:transcoder_loss_curves}. 

\subsection{Impact of Training Transcoders on Instruction-Tuning Data}
Since our base LLM used is an instruct model, a natural hypothesis is that transcoders fine-tuned on instruction-following data might learn features more relevant to CoT reasoning, thereby improving verification performance. To test this, we trained an version of our transcoders with instruction-tuning (IT) data. Starting from our pre-trained base transcoders, we continued fine-tuning for 1 epoch on the LMSYS-Chat-1M dataset~\citep{zheng2024lmsyschatm}, using the same hyperparameters as for the base transcoder training. Following the methodology of~\cite{lieberum-etal-2024-gemma}, we prepended and appended the Llama 3.1 8B Instruct model's IT prefixes to the user queries and model responses respectively.

However, as shown in Table~\ref{tab:base_it_transcoder_compare}, this additional training on IT data did not yield a consistent or meaningful improvement in verification performance on our tasks. This finding is consistent with recent work by~\cite{kissane2024saes}, who found that SAEs trained on base model activations can also faithfully reconstruct the activations of derived IT models. While a deeper mechanistic investigation into how instruction-tuning affects the underlying feature space is a promising direction, we leave this for future work. For our main experiments, we therefore use the more general base transcoders.

\begin{table}[t]
\centering
\resizebox{\textwidth}{!}{%
\begin{tabular}{@{}l ccc ccc ccc@{}}
\toprule
\multirow{2}{*}{\shortstack{{\bf Transcoder}\\{\bf Training}}} & \multicolumn{3}{c}{{\bf Synthetic (Boolean)}} & \multicolumn{3}{c}{{\bf Synthetic (Arithmetic)}} & \multicolumn{3}{c}{{\bf GSM8K}} \\
\cmidrule(l){2-4} \cmidrule(l){5-7} \cmidrule(l){8-10} 
 & AUROC $\uparrow$ & AUPR $\uparrow$ & FPR@95 $\downarrow$ & AUROC $\uparrow$ & AUPR $\uparrow$ & FPR@95 $\downarrow$ & AUROC $\uparrow$ & AUPR $\uparrow$ & FPR@95 $\downarrow$ \\ \midrule

 Base & 75.87 & 0.97 & 79.17 & 92.47 & 28.92 & 37.09 & 70.17 & 14.3 & 79.61 \\ 
 + IT Data & 76.04 & 1.20 & 66.82 & 91.39 & 28.44 & 38.47 & 72.01 & 15.40 & 83.27 \\ \bottomrule
\end{tabular}
}
\caption{Performance comparison of CRV with Base transcoders vs. transcoders further trained on Instruction-Tuning (IT) data. Arrows indicate preferred direction ($\uparrow$ higher is better, $\downarrow$ lower is better).}
\label{tab:base_it_transcoder_compare}
\vspace{0ex}
\end{table}

\subsection{Attribution Graph Computation}
\label{sec:app_attribution_graph}

{\bf Implementation Details.} We use the implementation from \cite{circuit-tracer} to compute attribution graphs. The primary hyperparameters were set as follows: a maximum of 4096 feature nodes, attribution traced from a maximum of 10 logit nodes (selected by a cumulative probability threshold of 0.95), and a batch size of 16 for backward passes. All other parameters follow the repository defaults.

\begin{table}[h]
\centering
\resizebox{\textwidth}{!}{%
\begin{tabular}{@{}l ccc ccc ccc@{}}
\toprule
\multirow{2}{*}{{\bf Attribution Position}} & \multicolumn{3}{c}{{\bf Synthetic (Boolean)}} & \multicolumn{3}{c}{{\bf Synthetic (Arithmetic)}} & \multicolumn{3}{c}{{\bf GSM8K}} \\
\cmidrule(l){2-4} \cmidrule(l){5-7} \cmidrule(l){8-10} 
 & AUROC $\uparrow$ & AUPR $\uparrow$ & FPR@95 $\downarrow$ & AUROC $\uparrow$ & AUPR $\uparrow$ & FPR@95 $\downarrow$ & AUROC $\uparrow$ & AUPR $\uparrow$ & FPR@95 $\downarrow$ \\ \midrule

 Before & 68.66 & 1.80 & 77.44 & 85.95 & 12.05 & 47.89 & 70.32 & 16.19 & 85.29 \\ 
 After & 75.87 & 0.97 & 79.17 & 92.47 & 28.92 & 37.09 & 70.17 & 14.3 & 79.61 \\ \bottomrule
\end{tabular}
}
\caption{Performance comparison of CRV using different token positions for attribution graph computation. The ``After'' setting computes the graph at the final token of the current step, while ``Before'' uses the final token of the previous step. Arrows indicate preferred direction ($\uparrow$ higher is better, $\downarrow$ lower is better).}
\label{tab:attribution_graph_results_compare}
\vspace{0ex}
\end{table}

\paragraph{\bf Ablation on Attribution Position.}
The attribution graph is computed with respect to a specific token position. The choice of this position is a critical methodological decision, as it determines which computational moment we analyze. We investigate two hypotheses: analyzing the state before a step is generated (the ``pre-computation'' trace) versus the state after it is complete (the ``post-computation'' trace). To test this, we compare two settings: (1) Before: computing the graph at the position of the final token of the previous reasoning step. For the first step of the CoT, this corresponds to the final token of the input question. (2) After: computing the graph at the final token of the current reasoning step, which is the default setting for our main experiments.

The results, presented in Table~\ref{tab:attribution_graph_results_compare}, show a clear and consistent advantage for the ``After'' setting across nearly all metrics and domains. We hypothesize that this is because the structural signatures of a flawed computation are most fully consolidated in the final token's representation after the step has been fully executed. The pre-computation state may contain signals of intent or planning, but the post-computation state contains the definitive trace of the executed algorithm, including the evidence of its failure. Based on these results, all experiments in the main body of the paper use the ``After'' (current step) position.

\clearpage

\section{Additional Classification Details}
\subsection{Attribution Graph Features}
\label{app:attribution_feature}
Here we give details about the extracted features for our attribution graphs that we used for our classifier. The feature set is organized into three hierarchical levels:

\paragraph{\bf 1. Global Graph Statistics:} These features provide a high-level summary of the pruned computational graph.
\begin{itemize}
    \item \textit{Node Counts:} The total number of active transcoder features, as well as the count of transcoder feature nodes and residual stream nodes remaining after pruning. This captures the overall sparsity and composition of the influential subgraph.
    \item \textit{Logit Statistics:} The probability of the top-ranked token and the entropy of the final logit distribution. These classic uncertainty measures serve as simple but informative baseline features.
\end{itemize}

\paragraph{\bf 2. Node Influence and Activation Statistics:} This group of features characterizes the properties of the nodes within the pruned graph, moving beyond simple counts.
\begin{itemize}
    \item \textit{Influence Scores:} The mean influence of all nodes in the pruned graph, along with the total and mean influence specifically from the residual stream (``error'') nodes. This helps quantify how much of the final output is attributed to specific learned features versus the model's direct pass-through states.
    \item \textit{Activation Statistics:} For the pruned transcoder feature nodes, we compute the mean, max, and standard deviation of their activation values. This captures the intensity and distribution of the active, interpretable features. A high maximum activation, for instance, might signal that a single, highly decisive feature was responsible for the step.
    \item \textit{Layer-wise Feature Histogram:} A histogram of active transcoder features across the model's layers. This feature vector characterizes the distribution of computational effort across the model's depth, allowing us to test hypotheses such as whether errors correlate with the activation of components at specific layers.
\end{itemize}

\paragraph{\bf 3. Topological and Path-Based Features:} To capture the structure and efficiency of the information flow, we compute a rich set of topological features on the pruned, directed subgraph.
\begin{itemize}
    \item \textit{Edge and Density Statistics:} Aggregate statistics on the edge weights (sum, mean, std), the total number of edges, and the graph density. We hypothesize that a sparse, fragmented graph (low density, few edges) may indicate a breakdown in information flow characteristic of an error.
    \item \textit{Centrality Measures:} To identify critical ``hub'' nodes in the computation, we calculate the mean and max for both degree centrality and weighted betweenness centrality. These features assess whether influence is concentrated or diffused.
    \item \textit{Connectivity and Path Lengths:} The number of weakly connected components and the average shortest path length within the largest component. A highly fragmented graph may suggest a failed computation. A particularly crucial feature is the {\bf shortest path length from any input token node to any final logit node}. This directly measures how efficiently information from the prompt propagates to the final decision. A long or non-existent path is hypothesized to be a strong signal that the model is ``ignoring'' its instructions or context.
\end{itemize}

\subsection{Additional Details on Baselines}
\label{app:baseline_classifiers}
Here we provide additional implementation details for the baseline methods used in our main experiments, ensuring full reproducibility.

\paragraph{\bf Black-Box Baselines.}
This category includes methods that operate solely on the output logits of the final token for each reasoning step. We use implementations from~\cite{wang2025latent}. 

\paragraph{\bf Gray-Box Baselines.}
This category includes methods that leverage the model's internal hidden states. For {\bf CoE}~\citep{wang2025latent} and {\bf CoT-Kinetics}~\citep{bi2025cot}, which are training-free, we followed the official implementations and protocols described by their respective authors to compute the verification scores. We set $\gamma$ in CoT-Kinetics to 0.8, and use mean pooling for reasoning token aggregation. 

For our supervised {\bf LR Probe} baseline, the choice of which layer's hidden states to use is a hyperparameter. To determine the optimal layer for each dataset, we performed a hyperparameter search, training a separate probe on the average hidden states from each of the 32 layers of Llama 3.1 8B Instruct on a small validation split. This process allowed us to identify the layer that contained the most predictive signal for each distinct reasoning task. The best-performing layers, which were subsequently used for the main results reported in Table~\ref{tab:main_results}, were found to be:
\begin{itemize}
    \item {\bf Layer 0} (the token embedding layer) for the Synthetic (Boolean) dataset.
    \item {\bf Layer 9} for the Synthetic (Arithmetic) dataset.
    \item {\bf Layer 0} (the token embedding layer) for the GSM8K dataset.
\end{itemize}

Similar to the protocol for the LR probe, we trained separate \textbf{MLP Probes} on the hidden states of specific layers. We searched over a hyperparameter grid similar to \cite{zhang2025reasoning}, varying the learning rate, weight decay, hidden dimension size ($d$), and the source layer index. The complete search space is provided in Table~\ref{tab:mlp_probe_param_search}. We selected the best hyperparameters for each dataset based on a small validation set. The optimal configurations are reported in Table~\ref{tab:mlp_probe_best_param}.

 \begin{table}[h]
    \centering
    \caption{Hyperparameter search space for the MLP Probe baseline.}
    \label{tab:mlp_probe_param_search}
    \begin{tabular}{@{}ll@{}}
    \toprule
    \textbf{Hyperparameter} & \textbf{Values Searched} \\ 
    \midrule
    Source Layer Index & $\{0, 1, \dots, 31\}$ \\
    Hidden Dimension ($d$) & $\{16, 32, 64, 128\}$ \\
    Learning Rate & $\{1\text{e-}3, 1\text{e-}4, 1\text{e-}5\}$ \\
    Weight Decay & $\{1\text{e-}1, 1\text{e-}2, 1\text{e-}3, 1\text{e-}4, 1\text{e-}5\}$ \\
    \bottomrule
    \end{tabular}
\end{table}

\begin{table}[h]
\centering
    \centering
    \caption{Optimal hyperparameters for the MLP Probe on each dataset. The Layer column indicates the hidden state layer that yielded the highest verification performance.}
    \label{tab:mlp_probe_best_param}
    \begin{tabular}{@{}lcccc@{}}
    \toprule
    \textbf{Dataset} & \textbf{Layer} & \textbf{Learning Rate} & \textbf{Weight Decay} & \textbf{Hidden Dim ($d$)} \\ 
    \midrule
    Synthetic (Boolean) & 0 & $1\text{e-}3$ & 1\text{e-}1 & 16 \\
    Synthetic (Arithmetic) & 12 & $1\text{e-}5$ & 1\text{e-}1 & 16 \\
    GSM8K & 0 & $1\text{e-}4$ & 1\text{e-}5 & 128 \\
    \bottomrule
    \end{tabular}
\end{table}

\subsection{Additional Details on Evaluation Metrics}
\label{app:metric_definitions}

We report three evaluation metrics: the Area Under the Receiver Operating Characteristic curve (AUROC), the False Positive Rate at 95\% True Positive Rate (FPR@95), and the Area Under the Precision–Recall curve (AUPR). For all metrics, we treat the verifier output as a real-valued score $s(x) \in \mathbb{R}$, where higher values indicate a higher predicted likelihood of an error. The ground-truth labels are $y \in \{0,1\}$, where $y=1$ denotes an incorrect reasoning step (the positive class) and $y=0$ denotes a correct step.

\textbf{AUROC.} The Area Under the ROC Curve is defined as the integral of the True Positive Rate (TPR) against the False Positive Rate (FPR):
\[
\text{AUROC} = \int_{0}^{1} \text{TPR}(\tau) \, d(\text{FPR}(\tau)).
\]
The rates at a given threshold $\tau$ are defined as:
\[
\text{TPR}(\tau) = \frac{\sum_x \mathbf{1}[s(x)\ge \tau \wedge y=1]}{\sum_x \mathbf{1}[y=1]}, 
\quad
\text{FPR}(\tau) = \frac{\sum_x \mathbf{1}[s(x)\ge \tau \wedge y=0]}{\sum_x \mathbf{1}[y=0]}.
\]
We compute this via the trapezoidal rule by sweeping $\tau$ across all observed score values.

\textbf{FPR@95.} This metric captures the False Positive Rate at a specific operating point where the recall (TPR) is fixed at 0.95. It is defined as:
\[
\text{FPR@95} = \text{FPR}(\tau_{0.95}),
\]
where $\tau_{0.95}$ is the threshold such that $\text{TPR}(\tau_{0.95}) = 0.95$. In practice, we select the strict threshold $\tau$ that yields a TPR closest to, but not less than, 0.95.

\textbf{AUPR.} The Area Under the Precision-Recall Curve quantifies the trade-off between precision and recall for the positive (error) class. It is defined as:
\[
\text{AUPR} = \int_{0}^{1} \text{Precision}(\text{Recall}) \, d(\text{Recall}),
\]
where Precision at threshold $\tau$ is:
\[
\text{Precision}(\tau) = \frac{\sum_x \mathbf{1}[s(x)\ge \tau \wedge y=1]}{\sum_x \mathbf{1}[s(x)\ge \tau]}.
\]
All metrics were computed using the standard implementations provided by the scikit-learn library~\citep{scikit-learn}, applied to the held-out test sets.

\begin{table}[h]
\centering
\resizebox{\textwidth}{!}{%
\begin{tabular}{@{}l ccc ccc ccc@{}}
\toprule
\multirow{2}{*}{{\bf Method}} & \multicolumn{3}{c}{{\bf Synthetic (Boolean)}} & \multicolumn{3}{c}{{\bf Synthetic (Arithmetic)}} & \multicolumn{3}{c}{{\bf GSM8K}} \\
\cmidrule(l){2-4} \cmidrule(l){5-7} \cmidrule(l){8-10} 
 & AUROC $\uparrow$ & AUPR $\uparrow$ & FPR@95 $\downarrow$ & AUROC $\uparrow$ & AUPR $\uparrow$ & FPR@95 $\downarrow$ & AUROC $\uparrow$ & AUPR $\uparrow$ & FPR@95 $\downarrow$ \\ \midrule
  Dummy & 50.8 & 0.25 & 100 & 49.84 & 1.20 & 100 & 48.06 & 6.46 & 100 \\ 
 Logistic Regression & 76.4 & 0.75 & 68.91 & 89.5 & 11.46 & 41.56 & 73.8 & 18.70 & 78.69 \\ 
 Random Forest & 61.71 & 4.49 & 100 & 92.99 & 43.68 & 30.56 & 71.7 & 17.65 & 76.18 \\ 
 Gradient Boosting & {75.87} & {0.97} & {79.17} & {92.47} & {28.92} & {37.09} & {70.17} & {14.3} & {79.61} \\ \bottomrule
\end{tabular}
}
\caption{Performance comparison of different diagnostic classifiers. Arrows indicate preferred direction ($\uparrow$ higher is better, $\downarrow$ lower is better).}
\label{tab:different_classifier_results}
\end{table}

\subsection{Additional Classifier and Their Results}
\label{app:appendix_classifiers}
To validate our choice of a Gradient Boosting classifier for the main experiments, we benchmarked its performance against several standard alternatives on our curated graph feature set. We evaluated a simple baseline, a linear model, and another tree-based ensemble to understand the trade-offs between model complexity and verification performance. For this analysis and main experiments in this work, we used the default hyperparameters from the scikit-learn library for each classifier, as an initial, non-exhaustive hyperparameter search did not yield any significant improvements, suggesting that the feature set itself provides a strong signal that is not overly sensitive to classifier configuration.

The results are presented in Table~\ref{tab:different_classifier_results}. As expected, the Dummy classifier, which makes predictions based on the training set's class distribution, performs near chance level (AUROC $\approx$ 50). This confirms that our graph features contain a significant predictive signal that is non-trivial to learn. Interestingly, a standard Logistic Regression model achieves competitive performance, yielding the best AUROC on two of the three datasets and the strongest overall results on GSM8K. This indicates that the features are highly informative even with a simple linear model.

However, the tree-based ensembles often achieve superior performance on other key metrics. The Random Forest classifier, for instance, yields a substantially higher AUPR and lower FPR@95 on the complex Arithmetic dataset, suggesting its ability to capture non-linear feature interactions is critical for high-precision verification in that domain. Overall, no single classifier is dominant across all domains and metrics. We chose Gradient Boosting for our main experiments as it consistently provides a strong and robust performance profile, but these results highlight that the optimal choice of diagnostic classifier may be domain-specific.

\subsection{Additional Results for RQs}
\label{app:additional_results}
\begin{figure}[t!]
    \centering
    \begin{subfigure}[b]{0.19\textwidth}
        \centering
        \includegraphics[width=\linewidth]{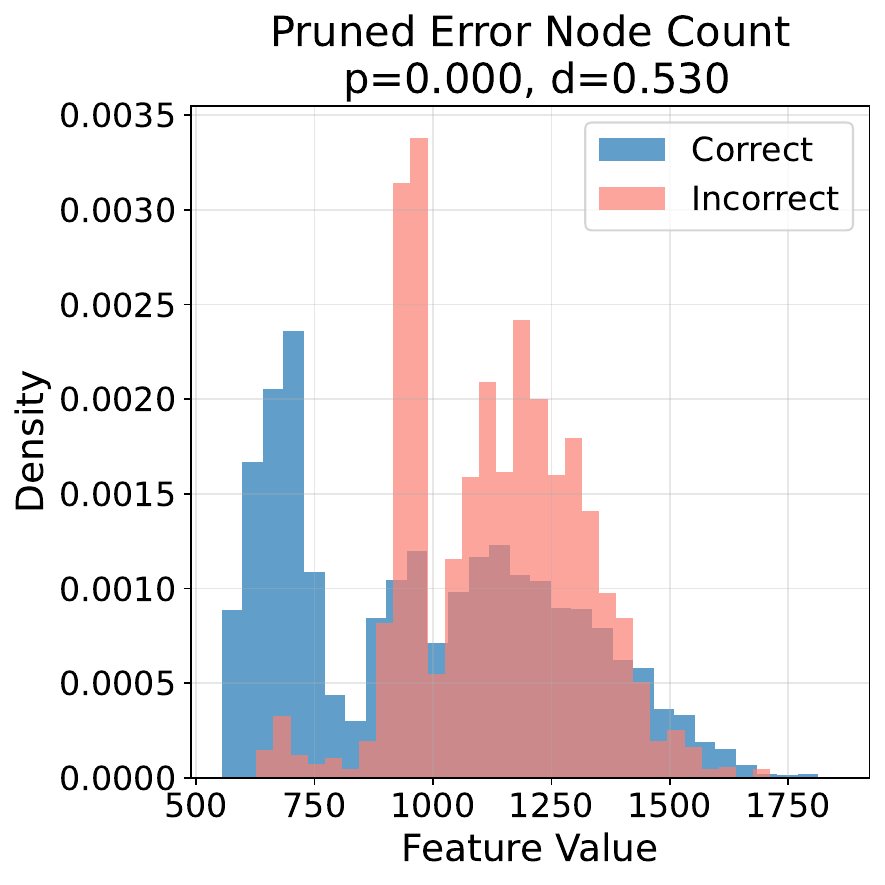} 
    \end{subfigure}
    \hfill
    \begin{subfigure}[b]{0.19\textwidth}
        \centering
        \includegraphics[width=\linewidth]{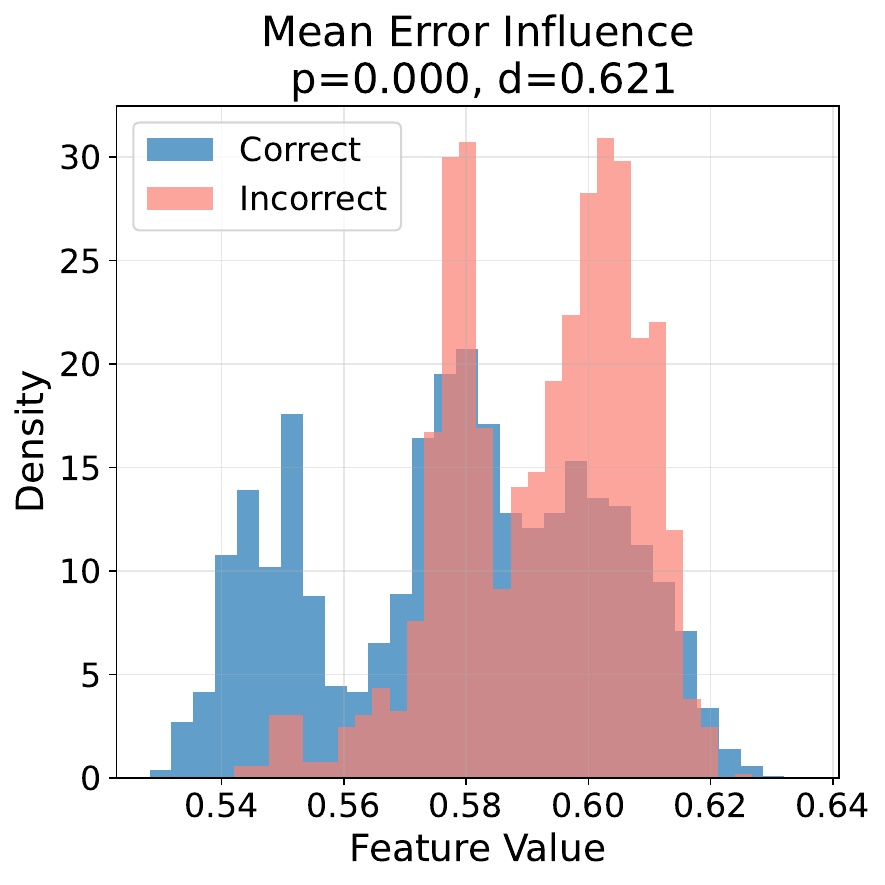} 
    \end{subfigure}
    \hfill
    \begin{subfigure}[b]{0.19\textwidth}
        \centering
        \includegraphics[width=\linewidth]{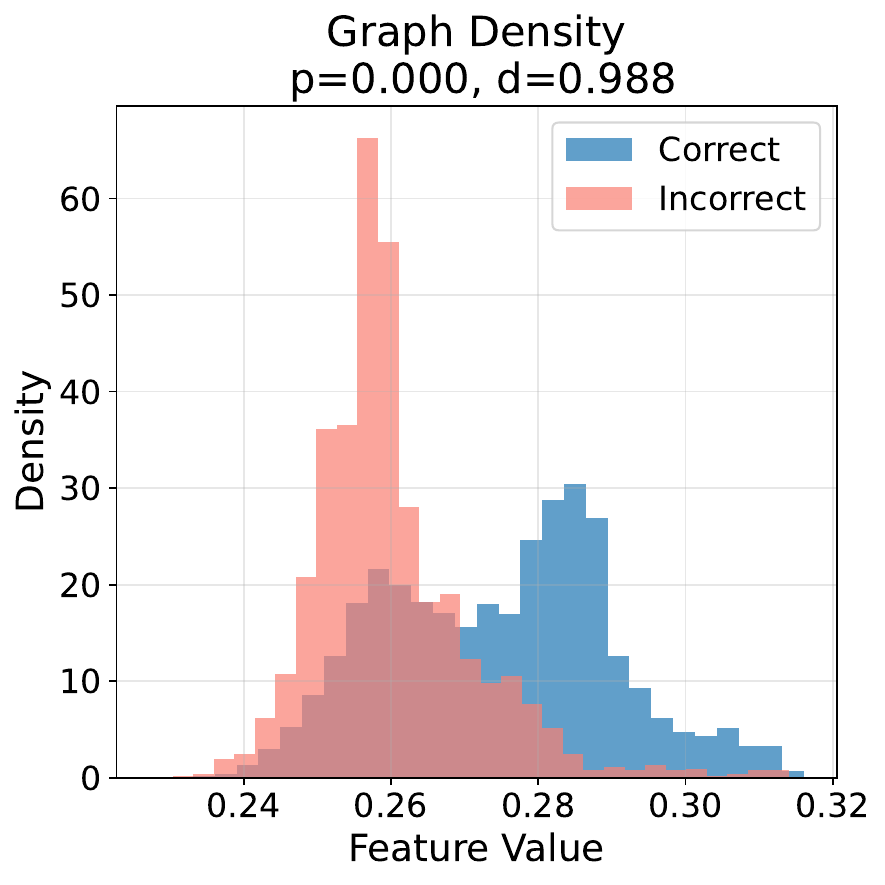} 
    \end{subfigure}
    \hfill
    \begin{subfigure}[b]{0.19\textwidth}
        \centering
        \includegraphics[width=\linewidth]{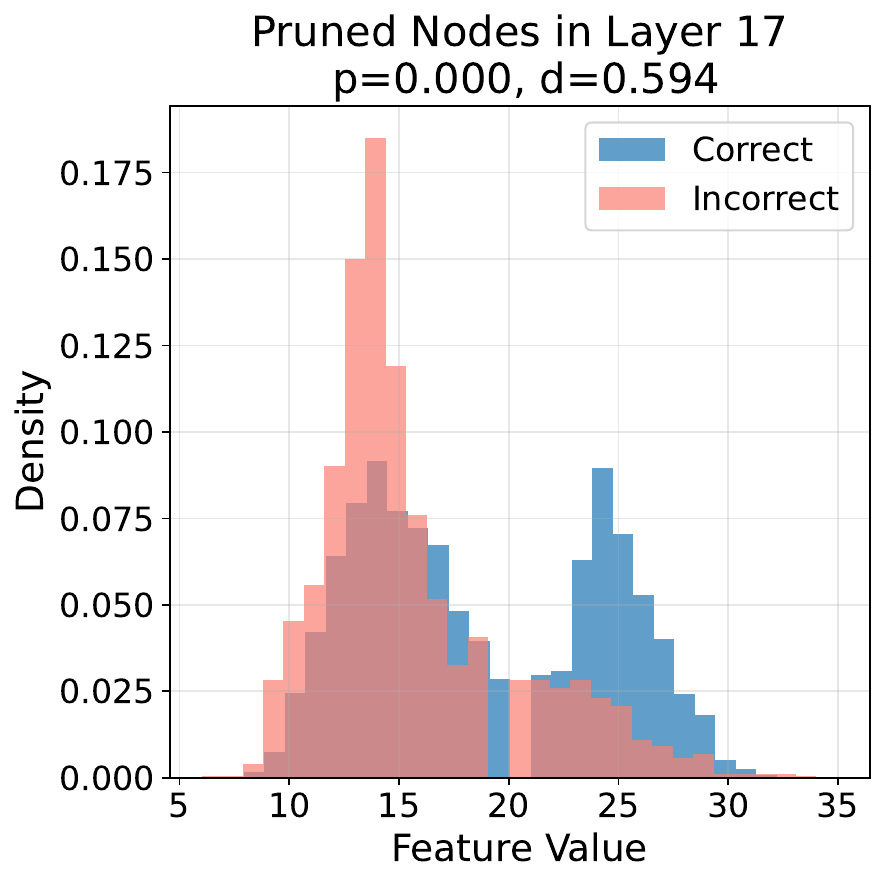} 
    \end{subfigure}
    \hfill
    \begin{subfigure}[b]{0.185\textwidth}
        \centering
        \includegraphics[width=\linewidth]{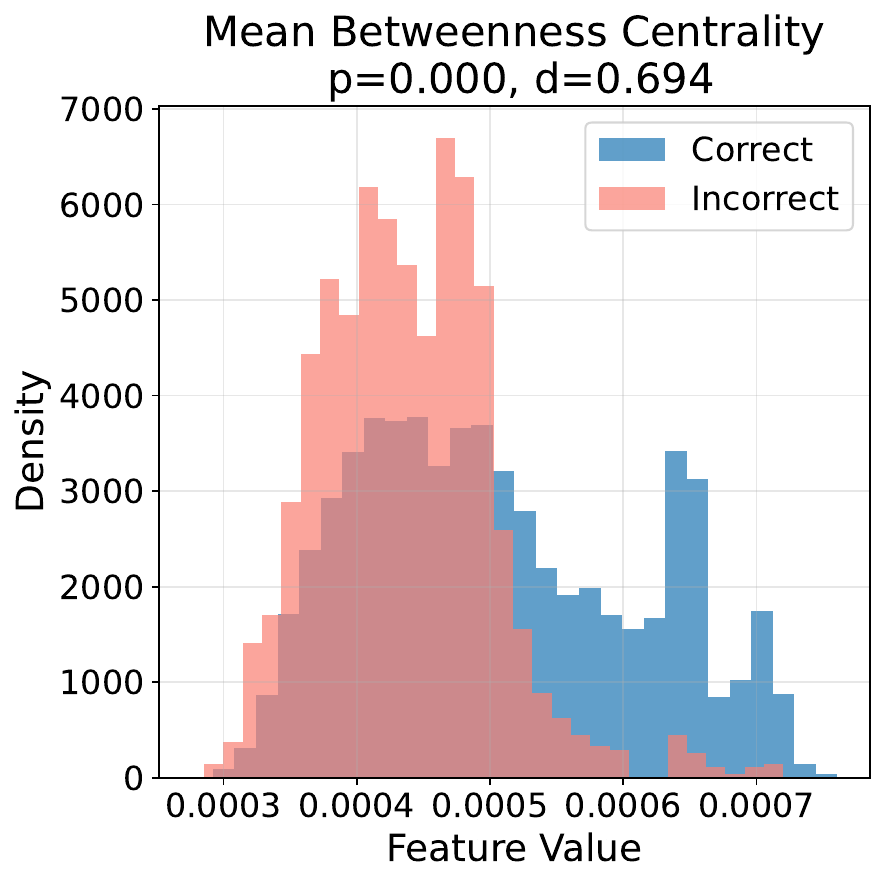} 
    \end{subfigure}
    \caption{
        {\bf Topological Fingerprints of Error on Arithmetic.} 
        Distributions of five selected graph features for correct (blue) vs. incorrect (red) reasoning steps. 
        The visual separation is statistically significant for each feature shown (independent t-test, $p < 0.001$) and represents a medium-to-large effect size (Cohen's d). This provides quantitative evidence that attribution graphs contain a clear, separable structural signal of a computation's integrity.
    }
    \label{fig:feature_distributions_arith}
\end{figure}

\begin{figure}[t!]
    \centering
    \begin{subfigure}[b]{0.19\textwidth}
        \centering
        \includegraphics[width=\linewidth]{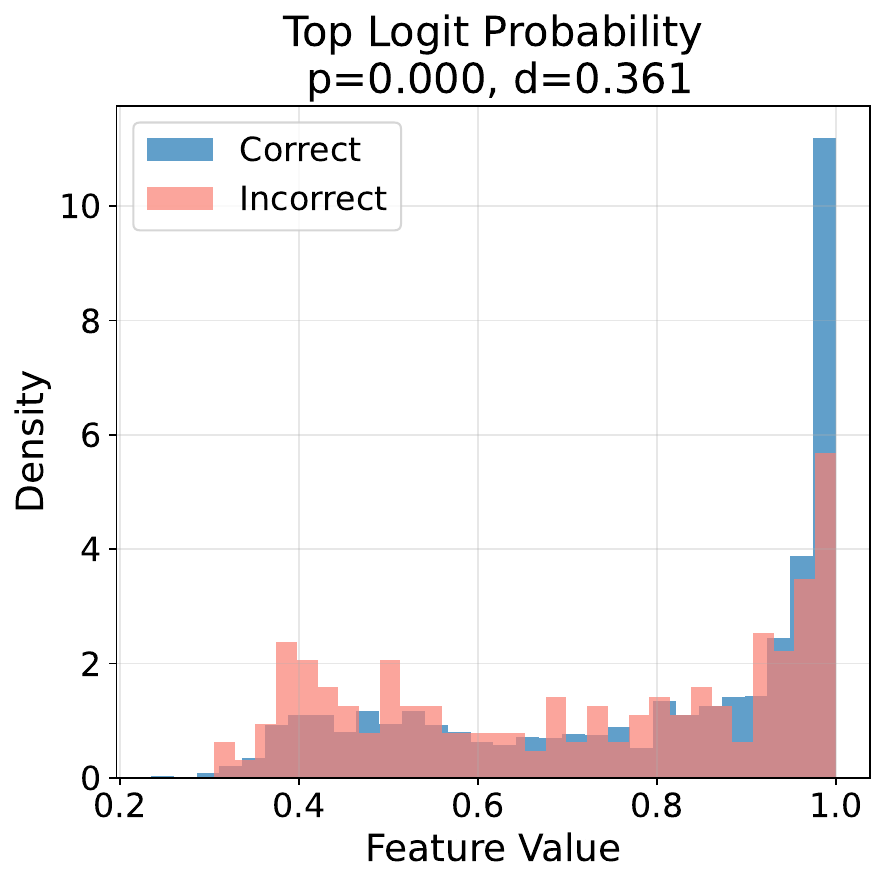} 
    \end{subfigure}
    \hfill
    \begin{subfigure}[b]{0.19\textwidth}
        \centering
        \includegraphics[width=\linewidth]{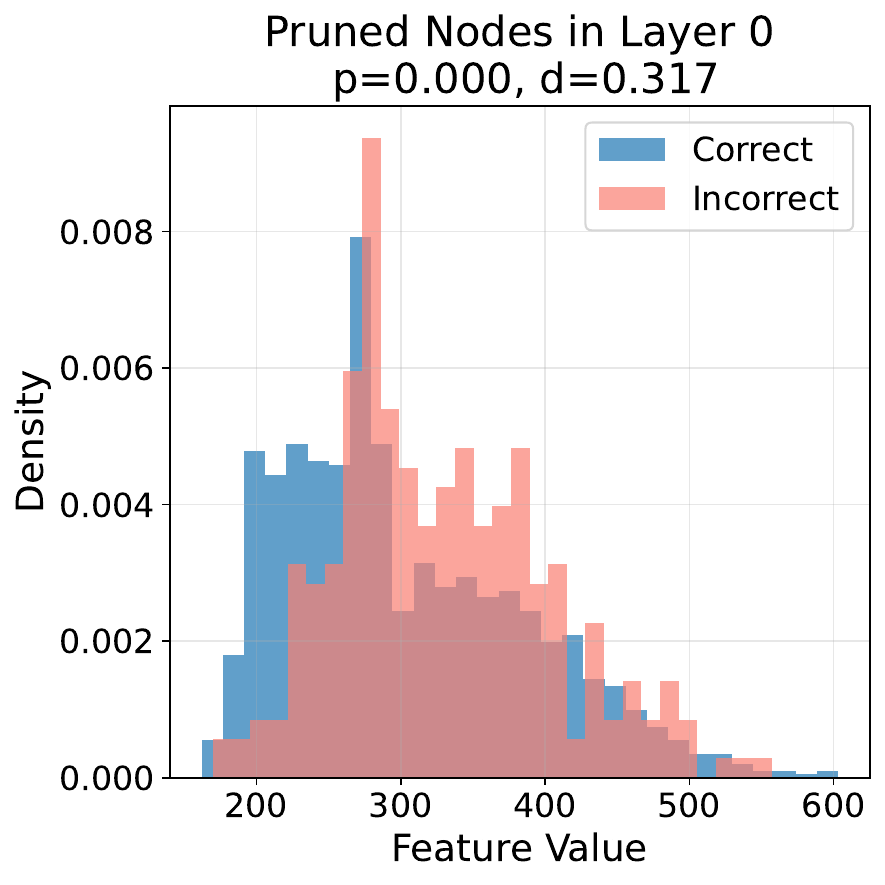} 
    \end{subfigure}
    \hfill
    \begin{subfigure}[b]{0.19\textwidth}
        \centering
        \includegraphics[width=\linewidth]{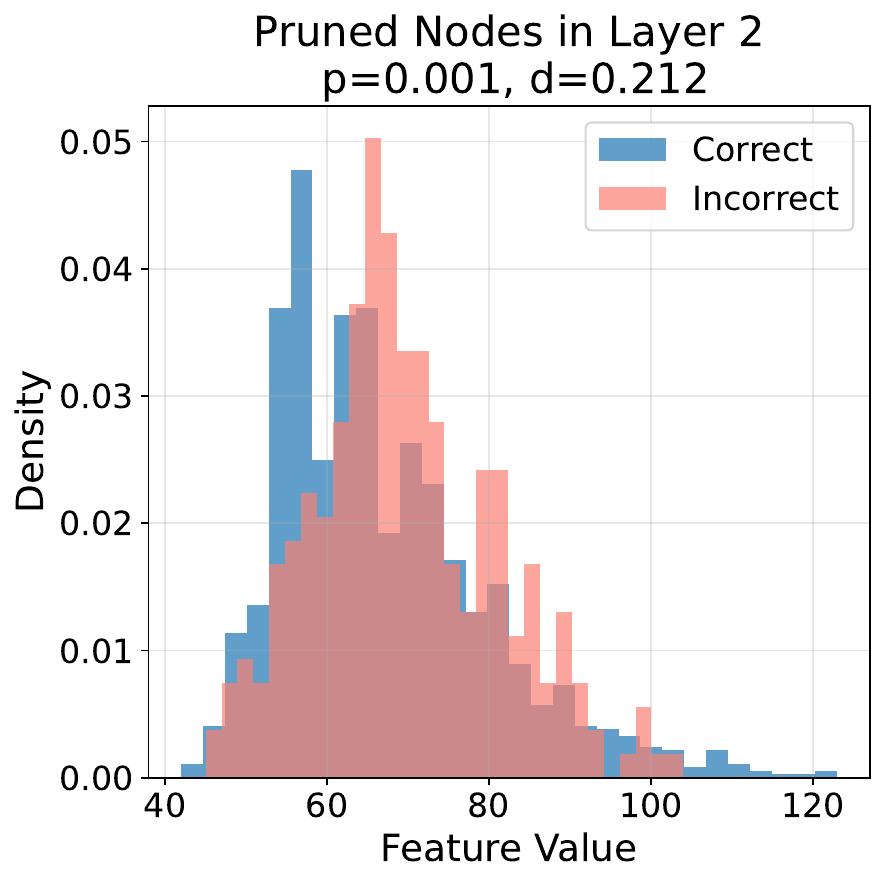} 
    \end{subfigure}
    \hfill
    \begin{subfigure}[b]{0.19\textwidth}
        \centering
        \includegraphics[width=\linewidth]{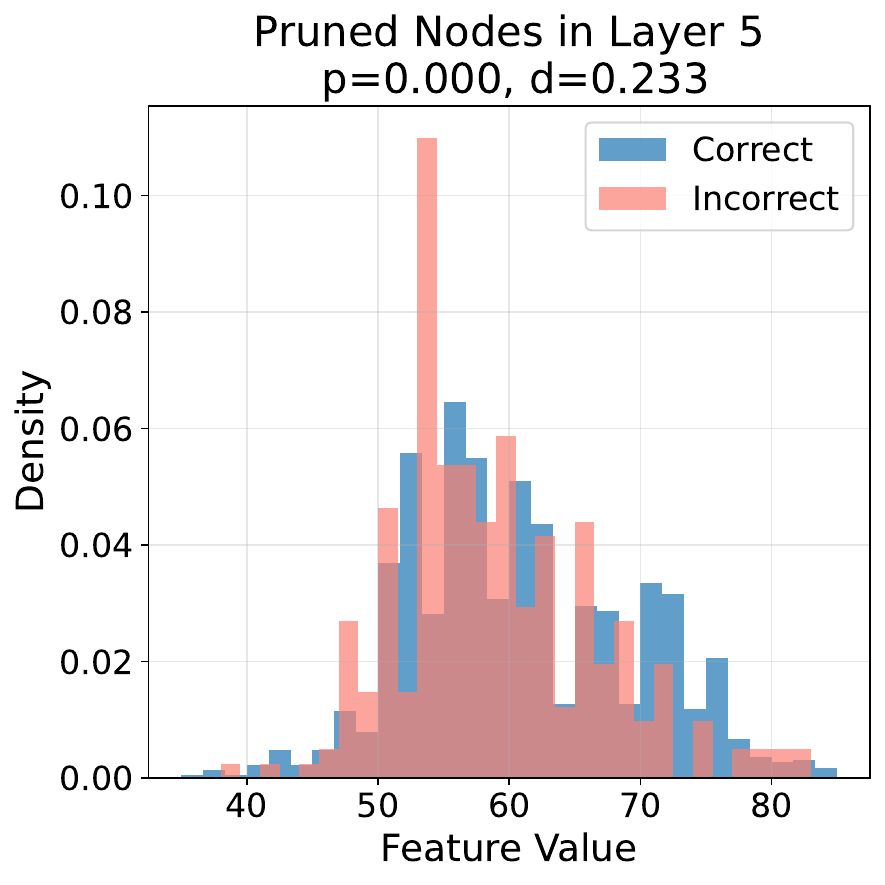} 
    \end{subfigure}
    \hfill
    \begin{subfigure}[b]{0.185\textwidth}
        \centering
        \includegraphics[width=\linewidth]{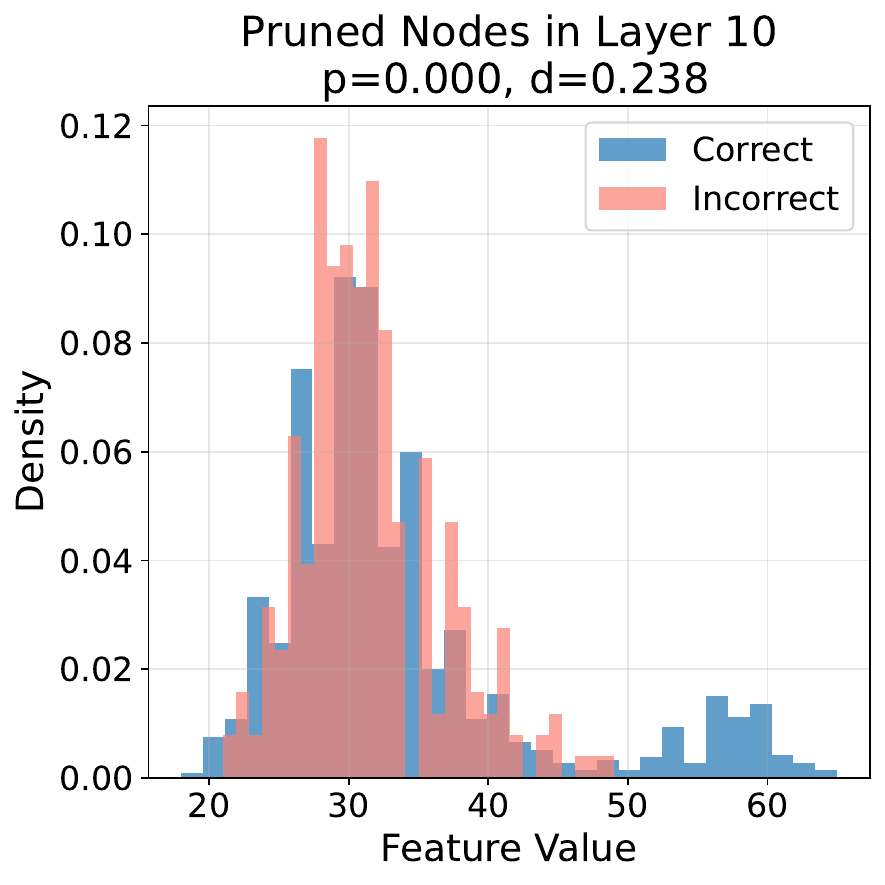} 
    \end{subfigure}
    \caption{
        {\bf Topological Fingerprints of Error on Boolean.} 
        Distributions of five selected graph features for correct (blue) vs. incorrect (red) reasoning steps. 
        The visual separation is statistically significant for each feature shown (independent t-test, $p < 0.001$) and represents a medium-to-large effect size (Cohen's d). This provides quantitative evidence that attribution graphs contain a clear, separable structural signal of a computation's integrity.
    }
    \label{fig:feature_distributions_bool}
\end{figure}

\begin{figure}
    \centering
    \begin{subfigure}[b]{0.32\linewidth}
        \centering
        \includegraphics[width=\linewidth]{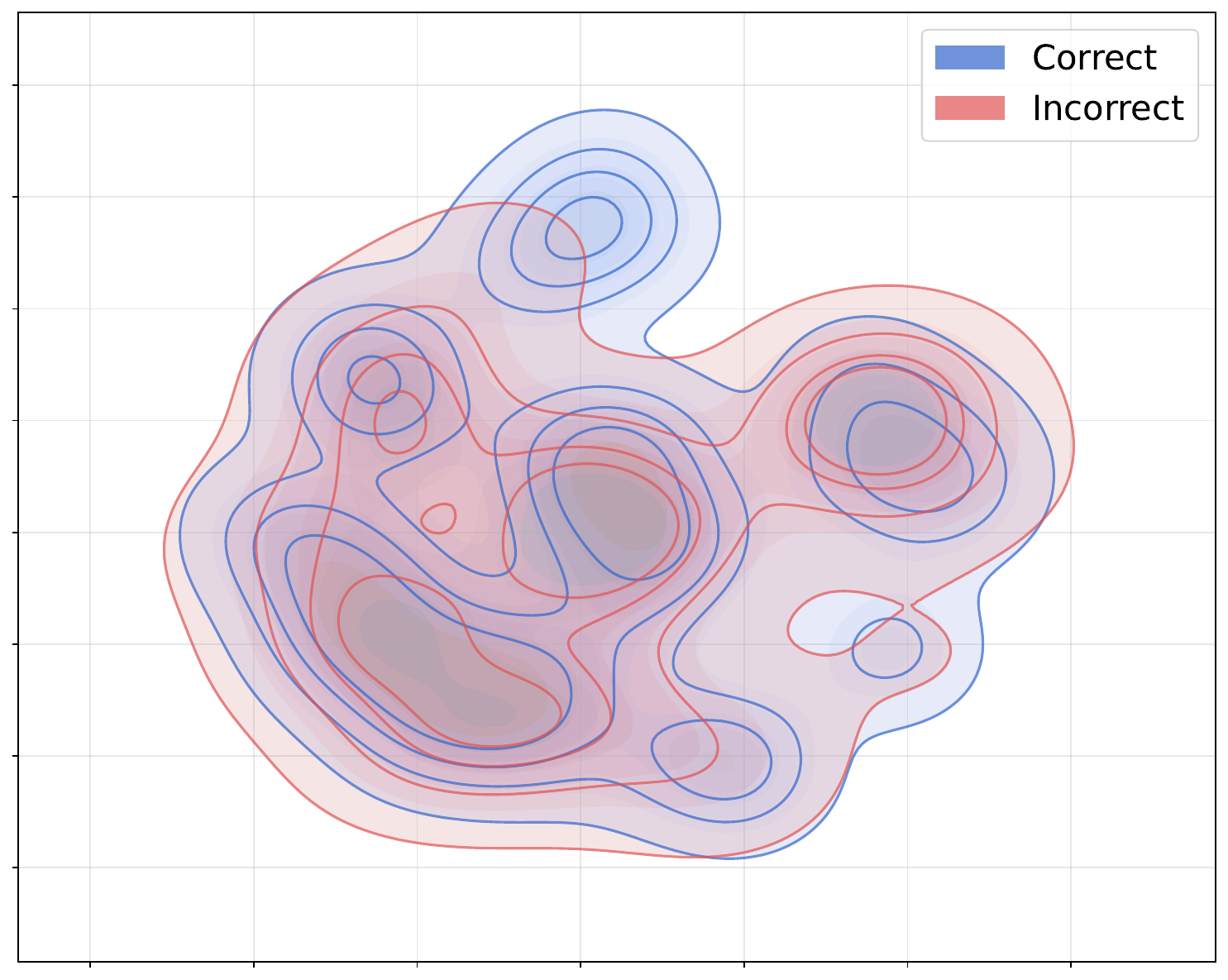} 
        \caption{Boolean}
    \end{subfigure}
    \hfill
    \begin{subfigure}[b]{0.32\linewidth}
        \centering
        \includegraphics[width=\linewidth]{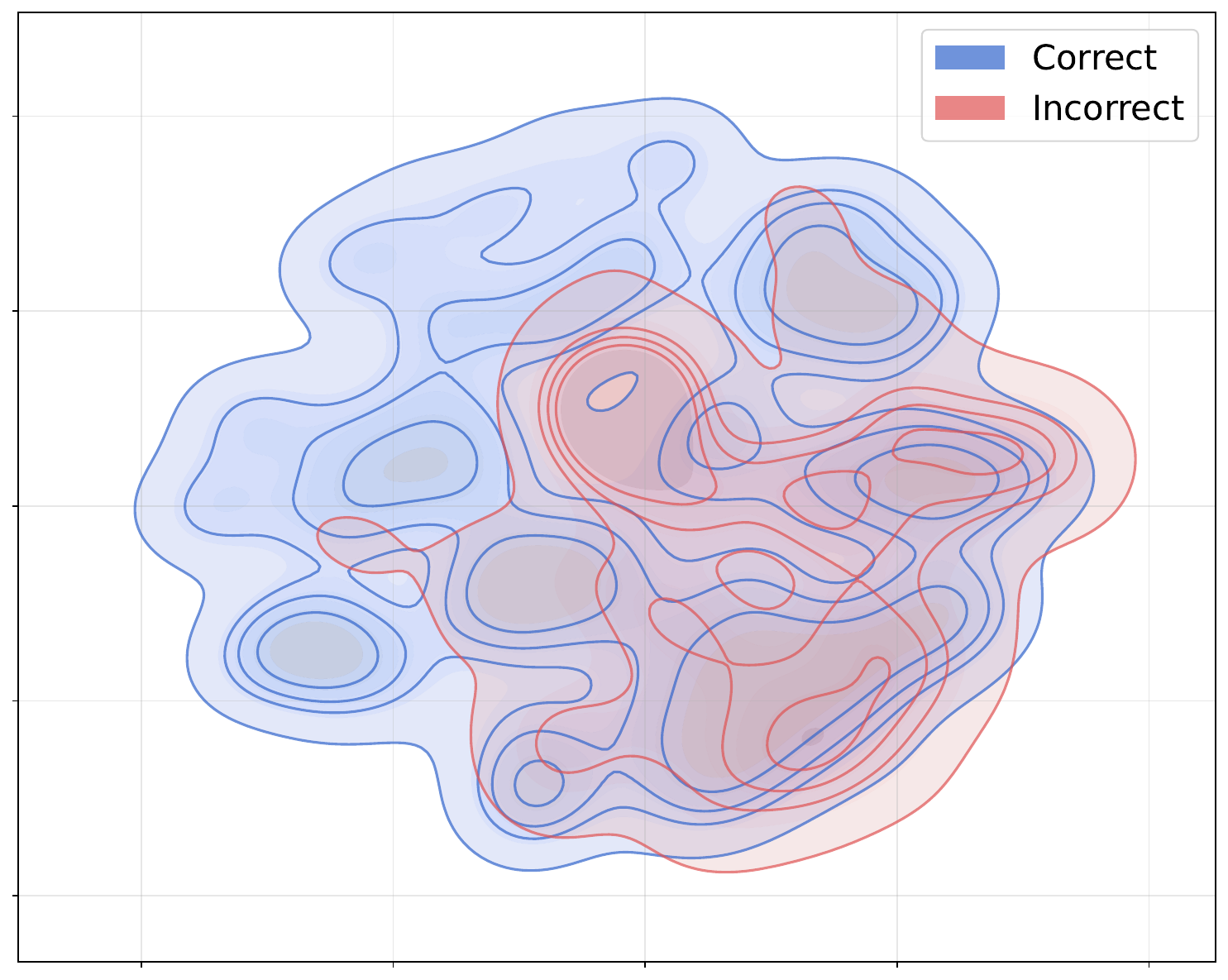} 
        \caption{Arithmetic}
    \end{subfigure}
    \hfill
    \begin{subfigure}[b]{0.32\linewidth}
        \centering
        \includegraphics[width=\linewidth]{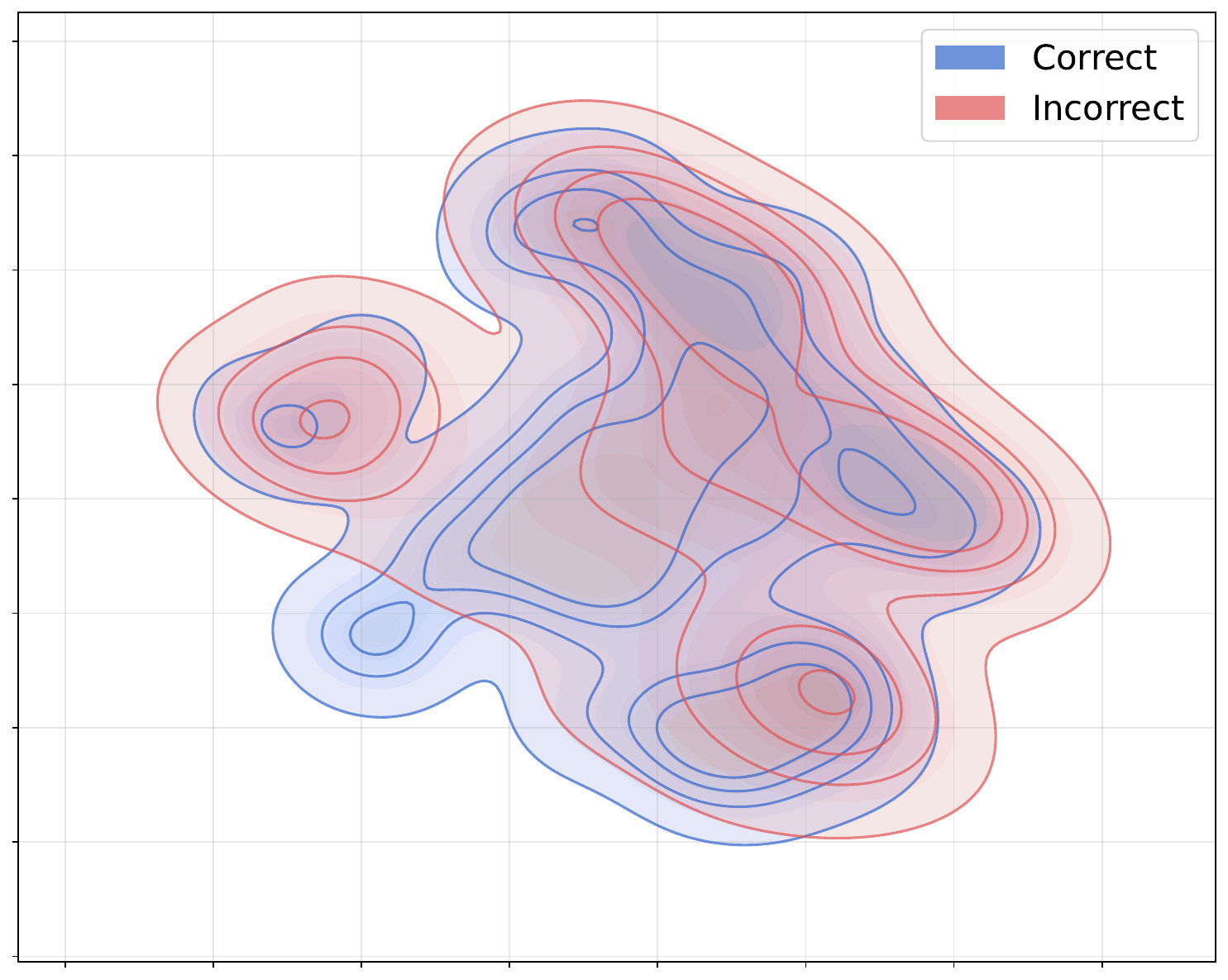} 
        \caption{GSM8K}
    \end{subfigure}
    \caption{
        Distributions of features after t-SNE for correct (blue) vs. incorrect (red) reasoning steps. The separation validates that CRV captures a distinct structural fingerprint of error.
    }
    \label{fig:dist_comparison_tsne}
\end{figure}

Here we provide additional results for our research questions. We first show distributions of highly predictive features for correct versus incorrect reasoning steps on our synthetic datasets (Figure~\ref{fig:feature_distributions_arith} for arithmetic; Figure~\ref{fig:feature_distributions_bool} for Boolean). Next, we display the distributions of full feature vectors after t-SNE projection in Figure~\ref{fig:dist_comparison_tsne}.

We demonstrate another casual intervention with a concrete case study on the arithmetic task, where we correct a subtle mathematical parsing error not by suppressing a faulty feature, but by \textit{amplifying} a correct one. We present the model with the expression \texttt{(-(5+(4*9)))}. As shown in Table~\ref{tab:intervention_example_1}, the model initially misinterprets the leading unary minus, treating it as a subtraction from an implicit -1 and producing the incorrect result of -42. Our CRV classifier flagged this step as incorrect, and its feature importance analysis suggested that the error stemmed from the absence of a strong ``multiplication'' signal.

Hypothesizing that the model failed to recognize the unary minus as a multiplication by -1, we performed a direct intervention. We identified a transcoder feature from the graph whose semantics corresponded to multiplication and, using a forward hook, manually amplified its activation value just before the model generated its erroneous step. The effect was immediate and successful. As shown in the ``After Intervention'' column, this boost was sufficient to steer the model onto the correct computational path. It now correctly multiplies by -1 and arrives at the final answer of -41. 

\begin{table}[h!]
\centering
\begin{tabular}{p{0.47\textwidth} p{0.47\textwidth}}
\toprule
{\bf Before Intervention (Incorrect)} & {\bf After Intervention (Correct)} \\
\midrule

\parbox[t]{\linewidth}{
    Evaluate the arithmetic expression below. \par
    ( - ( 5 + ( 4 * 9 ) ) ) \par
    To evaluate this expression, we need to follow the order of operations (PEMDAS): \par
    1. Multiply 4 and 9: 4 * 9 = 36 \par
    2. Add 5 and 36: 5 + 36 = 41 \par 
    3. \sethlcolor{myred}\hl{Subtract 41 from -1: -1 - 41 = -42} \par
    \sethlcolor{myred}\hl{The final answer is -42.}
} &

\parbox[t]{\linewidth}{
    Evaluate the arithmetic expression below. \par
    ( - ( 5 + ( 4 * 9 ) ) ) \par
    To evaluate this expression, we need to follow the order of operations (PEMDAS): \par
    1. Multiply 4 and 9: 4 * 9 = 36 \par
    2. Add 5 and 36: 5 + 36 = 41 \par 
    3. \sethlcolor{myblue}\hl{Multiply -1 by 41: -1 * 41 = -41} \par
    \sethlcolor{myblue}\hl{The final answer is: -41.} \par
} \\
\bottomrule
\end{tabular}
\caption{A successful causal intervention via feature amplification. A \sethlcolor{myred}\hl{subtraction} error is corrected by amplifying a \sethlcolor{myblue}\hl{multiplication} feature, steering the model to the correct reasoning path.}
\label{tab:intervention_example_1}
\end{table}

\clearpage

\section{Usage of Language Language Models}
In the preparation of this manuscript, we utilized LLMs as writing assistants. Their role was strictly limited to improving the clarity, conciseness, and grammatical correctness of the text. The authors take full responsibility for all content and any remaining errors.

\section{Limitations}

Our work introduces a new methodology for the scientific analysis of reasoning, and its limitations are intrinsically linked to its design as a white-box, mechanistic instrument.

\paragraph{\bf Computational Intensity.}
A primary limitation of CRV is its computational intensity. The process, which involves training a suite of transcoders, replacing model modules, and constructing a detailed attribution graph for every reasoning step, is orders of magnitude more resource-intensive than black-box or gray-box verification methods. This cost is a direct consequence of our white-box approach, which prioritizes mechanistic depth over practical efficiency. As such, CRV in its current form is positioned as a scientific tool for deep analysis, not as a scalable, real-time verifier for production systems.

\paragraph{\bf Aggregative vs. Feature-Level Analysis.}
The feature set used by CRV is primarily aggregative; it captures statistical and topological properties of the graph, such as node counts, influence scores, and density. As an early work, it does not yet fully exploit the semantic content of the individual transcoder features that constitute the graph's nodes. For instance, our current classifier learns statistical correlations over the entire feature set; it does not reason symbolically about whether a specific feature for numerical addition is appropriately activated by numerical inputs. This represents a significant opportunity. A promising future direction lies in developing more sophisticated classifiers or rule-based systems that operate directly on the semantics of these disentangled features, paving the way for a new class of neuro-symbolic verifiers.

\paragraph{\bf Generalizability of Error Signatures.}
Our empirical results are based on a single model family (Llama 3.1) at the 8B scale. Whether the precise structural fingerprints we identified generalize to different architectural paradigms, such as Mixture-of-Experts, or across significant model scales (e.g., 70B and larger) remains an open question. Furthermore, as our cross-domain experiments revealed, the error signatures are highly domain-specific. Our work provides a strong foundation and a methodology for discovering these signatures, but further studies are needed to determine if more universal principles of computational failure exist.

\paragraph{\bf Fidelity of Interpretability Tools.}
The validity of our analysis is contingent on the quality and fidelity of the underlying interpretability tools. The features identified by our transcoders, while demonstrably useful, represent one possible sparse basis and are not exhaustive. Similarly, the attribution method provides a powerful but ultimately incomplete approximation of the true information flow within the model. Future improvements in these foundational techniques, such as the development of more faithful sparse autoencoders or more precise attribution methods, will directly enhance the resolution and reliability of analyses like ours.

\end{document}